**Thermodynamic-Inspired Explainable GeoAI: Uncovering Regime-Dependent Mechanisms in Heterogeneous Spatial Systems**

Sooyoung Lim[1], Zhenlong Li[1*], Zi-Kui Liu[2]

[1]Geoinformation and Big Data Research Lab, Department of Geography, Pennsylvania State University, University Park, PA 16801, USA

[2]Department of Materials Science and Engineering, Pennsylvania State University, University Park, PA 16801, USA

[*]Correspondence: zhenlong@psu.edu

**Abstract**

Modeling spatial heterogeneity and associated critical transitions remains a fundamental challenge in geography and environmental science. While conventional Geographically Weighted Regression (GWR) and deep learning models have improved predictive skill, they often fail to elucidate state-dependent nonlinearities where the functional roles of drivers represent opposing effects across heterogeneous domains. We introduce a thermodynamics-inspired explainable geospatial AI framework that integrates statistical mechanics with graph neural networks. By conceptualizing spatial variability as a thermodynamic competition between system Burden ($E$) and Capacity ($S$), our model disentangles the latent mechanisms driving spatial processes. Using three simulation datasets and three real-word datasets across distinct domains (housing markets, mental health prevalence, and wildfire-induced PM2.5 anomalies), we show that the new framework successfully identifies regime-dependent role reversals of predictors that standard baselines miss. Notably, the framework explicitly diagnoses the phase transition into a Burden-dominated regime during the 2023 Canadian wildfire event, distinguishing physical mechanism shifts from statistical outliers. These findings demonstrate that thermodynamic constraints can improve the interpretability of GeoAI while preserving strong predictive performance in complex spatial systems.

## Introduction

Modeling spatial heterogeneity and associated critical transitions in complex spatial systems remain a fundamental challenge, because conventional models often struggle with state-dependent nonlinearities in which the functional roles of drivers shift across heterogeneous spatial domains (Reichstein et al., 2019; Jemeļjanova et al., 2024; Koldasbayeva et al., 2024). Whether examining the contagion of violence, the diffusion of infectious disease, or the dispersion of atmospheric pollutants, traditional spatial statistics have long established that assuming global uniformity leads to biased estimates and ineffective policy interventions (Anselin, 1989; Banerjee et al., 2003). While methods like Geographically Weighted Regression (GWR) attempt to capture these spatially varying relationships (Brunsdon et al., 1996; Fotheringham et al., 2009; Dambon et al., 2021), they rely on linear functional forms and pre-specified kernels that can be too restrictive for non-linear, high-dimensional interactions underlying many real-world spatial processes. Moreover, these models primarily provide a descriptive map of where coefficients vary, but often offer limited structural guidance on why they vary, and may absorb local noise as meaningful heterogeneity when the signal-to-noise ratio is low (Wheeler & Tiefelsdorf, 2005; Scheffer et al., 2009; Dakos et al., 2012).

Conversely, the recent advances in geospatial artificial intelligence (GeoAI) and deep learning have achieved superior predictive accuracy (Goodfellow et al., 2016; De Sabbata & Liu, 2023; Yin et al., 2024). However, these gains often rely on opaque latent representations that limit mechanistic interpretation and policy-relevant diagnosis, particularly when spatial processes are non-stationary (Goodfellow et al., 2016; Janowicz et al., 2020; Liu & Biljecki, 2022). A critical, often overlooked limitation is that the widely used explainable AI techniques, such as SHapley Additive exPlanations (SHAP) or Local Interpretable Model agnostic Explanations (LIME), are primarily post-hoc attribution tools that quantify feature contributions but do not explicitly separate opposing functional pathways (Ribeiro et al., 2016; Lundberg & Lee, 2017; Adadi & Berrada, 2018). While these methods can identify influential predictors, they do not explicitly decouple whether elevated outcomes arise from accumulated structural stressors (Burden) or from diminished adaptive buffering (Capacity), which is essential for hypothesis-driven diagnosis in geographic research (Rudin, 2019).

The limitation becomes more critical when predictor roles are not stable across space or system states. Many existing models either assume fixed functional roles of predictors (or smoothly varying local roles), which makes it difficult to represent regime-dependent role reversals, including sign changes or interaction-driven inversions. Similar regime-dependent sign reversals have been documented in ecological and climatic systems approaching tipping points, where drivers that are stabilizing under one regime become destabilizing as the system nears a critical threshold (Kéfi et al., 2013; Dakos et al., 2012; Scheffer et al., 2009). A variable acting as a protective mechanism in one context may transform into a risk amplifier in another, yet standard regression and black-box machine learning models lack an explicit structural prior to express this "role reversal" in a way that remains interpretable and comparable across space.

To bridge this gap, we introduce a thermodynamics-inspired GeoAI framework with an explicit structural prior. This framework views spatial outcomes as emergent from the competition between disorder-inducing Burden $E$ and order-maintaining Capacity $S$, summarized through a latent free-energy functional $F(x) = E(x) - TS(x)$, where $T$ is not a literal thermal metric, but a learnable effective temperature capturing the regime-specific stochasticity. We frame complex spatial systems as thermodynamic-like systems for modeling purposes; thus, a high $T$ implies a regime dominated by entropic fluctuations, whereas a low $T$ indicates a deterministic regime driven primarily by Burden potential. Drawing on Zentropy theory (Liu, 2024; Liu, 2025), we view each location as a thermodynamic-like ensemble state and explicitly learn burden-capacity decomposition during model training. This design aligns with physics-informed machine learning in that it embeds a structural constraint and prior into the learning process to improve robustness, generalization, and interpretability, while acknowledging that the learned components are latent constructs rather than direct physical measurements (Raissi et al., 2019; Karniadakis et al., 2021; Reichstein et al., 2019).

Crucially, this framework serves not only as a predictive model but as a diagnostic instrument for characterizing regime-dependent variable roles. It injects domain knowledge directly into model structure. Following standard practice in regression analysis, we specify a priori which variables are expected to increase outcome (burden-like) versus reduce it (capacity-like) and encode this expectation by assigning predictors to the E and S channels. We treat this grouping as soft inductive bias rather than a rigid physical constraint. This initialization transforms the pre-classification into a falsifiable hypothesis: the model retains the flexibility to diagnose role reversals (learning negative sensitivities for burden-assigned variables), when the data reveal opposing regime-dependent interactions. By examining the learned signed sensitivities of F(x), E(x), and S(x) across regimes and space, we extract interpretable patterns of state-dependent functional roles, moving beyond purely linear and stationary assumptions. In this way, our approach complements early-warning and regime-shift concepts by providing a structured decomposition that can surface state-dependent role changes in complex spatial systems (Scheffer et al., 2009; Dakos et al., 2012; Ma et al., 2025). Zentropy-based neural network (ZENN) has introduced free-energy style decomposition in generic neural architectures (Wang et al., 2026); our framework, named ZeGNN, embeds this decomposition into a spatial graph neural network (GNN) with regime-switching gating, enabling explicitly spatial, non-stationary burden-capacity diagnostics in geographic systems.

We validate the generalizability of this framework using three simulation datasets and three real-word datasets across distinct classes of complex spatial systems: housing price (a socioeconomic system), mental health (a public health system), and PM2.5 dynamics (a physical-environmental system). Despite their different generative mechanisms, we show that these systems occupy interpretable regions of a shared burden-capacity phase space, and that regime-conditioned decomposition provides a unified, associational diagnostic of where burden-like and capacity-like forces co-occur and dominate. Our findings suggest that embedding a thermodynamic structural prior into spatial deep learning can improve both predictive skill and interpretability, supporting geographically grounded hypothesis generation about heterogeneous mechanisms and potential leverage points.

## Results

### ZeGNN models spatial outcomes as an entropy-adjusted mixture of thermodynamic regimes

ZeGNN models heterogeneous spatial outcomes through a thermodynamic mixture-of-regimes architecture that is explicitly designed for mechanism-oriented interpretation (Fig. 1). Predictors are partitioned a priori into burden and capacity feature blocks based on domain knowledge, where burden variables encode exposure, hazard, or socioeconomic disadvantage and capacity variables capture adaptive resources, buffering infrastructure, or environmental ventilation (Chen et al., 2024; Clark et al., 2024). We treat this burden-capacity partition as a soft inductive bias rather than a hard sign constraint. The model then learns regime-specific burden and capacity potentials and combines them through graph-aware regime gating over a spatial neighborhood structure. The thermodynamic role of each predictor is evaluated post hoc from the learned signed responses of $F(x)$, $E(x)$, and $S(x)$, rather than being imposed a priori (Reichstein et al., 2019; Schölkopf et al., 2021; Liu, 2024). Throughout, $F, E$, and $S$ are interpreted as structured summaries of conditional associations rather than as causal effects. In this sense, ZeGNN builds on recent Zentropy-enhanced neural architectures that introduce free-energy decomposition into deep learning (Wang et al., 2026), but extends them to explicit spatial graphs and regime-switching parameterization for GeoAI. This formulation allows the predicted surface to be decomposed into burden-dominant and capacity-dominant contributions while permitting the governing relationships to vary across space, rather than imposing a single stationary response function over the entire study domain.

For spatial unit $i$ and latent regime $k$, the regime-specific free-energy potential $F_{ik}$ is defined as

$$F_{ik} = E_k\left(x_i^E\right) - T_k S_k\left(x_i^S\right), \quad T_k > 0 \tag{1}$$

Where $E_k(x_i^E)$ and $S_k\left(x_i^S\right)$ denote the regime-specific burden and capacity potentials learned from the burden and capacity feature blocks, respectively, and $T_k$ is a regime-specific effective temperature. The final standardized prediction is given by an entropy-adjusted mixture of regime free energies:

$$\hat{z}_i = F_i = \sum_{k=1}^{K} p_{ik} F_{ik} + T_{eff,i}\,\overline{H}_i \tag{2}$$

Where $p_{ik}$ is the soft membership probability of spatial unit $i$ in regime $k$, and $\overline{H}_i$ is the normalized gating entropy:

$$\overline{H}_i = -\frac{1}{\log K} \sum_{k=1}^{K} p_{ik} \log(p_{ik} + \epsilon) \tag{3}$$

and $T_{eff,i}$ is the mixture-average effective temperature:

$$T_{eff,i} = \sum_{k=1}^{K} p_{ik} T_k \tag{4}$$

For interpretation, we additionally define the mixture-averaged burden and capacity field as:

$$E_i = \sum_{k=1}^{K} p_{ik} E_k(x_i^E), \qquad S_i = \sum_{k=1}^{K} p_{ik} S_k(x_i^S). \tag{5}$$

Accordingly, the entropy term modifies the final free-energy prediction $F_i$, whereas $S_i$ remains the mixture-averaged capacity field rather than an entropy-corrected quantity. Spatial dependence enters the model through a graph-based gating mechanism defined on $k$-nearest-neighbor representation of the study domain. The gating network takes standardized covariates and coordinates as input, diffuses regime logits across the spatial graph, and outputs soft regime probabilities $p_{ik}$. This allows the model to recover geographically coherent regimes and transition zones instead of forcing a single global relationship. As illustrated in Fig. 1, ZeGNN therefore yields the predicted free energy $F$, its burden and capacity components $E$ and $S$, and the normalized uncertainty associated with regime competition.

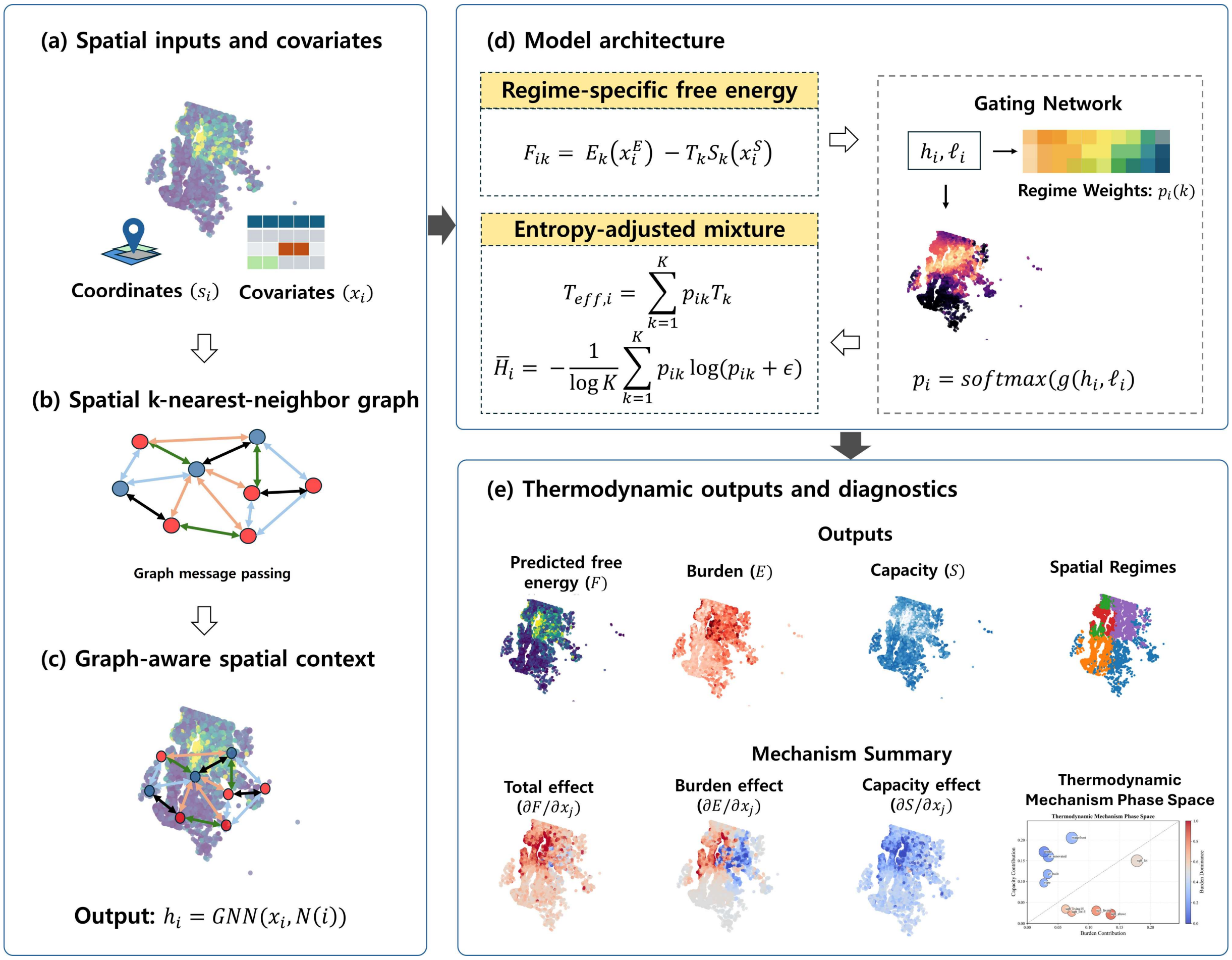

**Fig. 1. ZeGNN architecture and thermodynamic outputs:** (a): Spatial inputs and covariates defined over the study domain. (b): Construction of the spatial k-nearest-neighbor graph used to encode local spatial dependence. (c): Graph-aware spatial context, illustrating how neighborhood structure is overlaid on the study domain to propagate local information during gating. (d): ZeGNN model architecture, showing the regime-specific free-energy formulation and the entropy-adjusted mixture used for final prediction. (e): Thermodynamic decomposition, regime structure, and derived diagnostics from the fitted model.

## Evaluation with the simulation data

To verify that the proposed thermodynamic structural prior is not merely an interpretive overlay but a recoverable representation under controlled conditions, we first conducted a synthetic spatial experiment in which the ground-truth free energy and its decomposition are known. Specifically, we generated spatially heterogeneous outcomes from regime-dependent latent surfaces, allowing the true free energy $F$, burden $E$, and capacity $S$ to vary across space with clear regime boundaries. We considered three complementary simulation scenarios (Global Linear, Local Linear, and Non-Linear) that share the same spatial domain and covariate fields but differ in how the thermodynamic potential $E$ and $S$ are generated. Complete specifications and dataset overviews are provided in Supplementary Note 1 (Supplementary Fig. S1-S4). Fig. 2 summarizes how the ZeGNN recovers these latent thermodynamic components, identifies regimes, and internalizes spatial dependence relative to conventional baselines, including regime-dependent nonlinear response forms (Fig. 2A) that cannot be captured by smooth-coefficient variation alone.

The learned decomposition reproduces the large-scale structure of the ground-truth thermodynamic fields. The predicted free energy surface closely matches the true pattern, including the major sign transitions and coherent patches that arise from regime-wise mechanisms (Fig. 2C), and the model simultaneously reconstructs the burden field and the capacity field with spatial coherence (Figs. 2A, C), showing that the fitted outcome surface can be interpreted as an emergent balance between disorder-inducing and buffering pathways rather than an undifferentiated latent embedding. While $S$ is generally harder to learn because it is expressed through the entropic pathway and is modulated by regime mixing, the predicted $S$ field still preserves the dominant spatial gradients implied by the data-generating process, supporting that the decomposition is identifiable in practice under spatial heterogeneity.

This controlled setting also clarifies the generalization behavior under spatially structured data. When evaluated using both random folds and spatially blocked folds, the ZeGNN maintains strong predictive skill and exhibits reduced degradation under spatial blocking relative to non-spatial or weakly structured learners (Fig. 2D), consistent with the purpose of explicitly modeling spatial dependence and heterogeneity rather than leaving it in residual structure (Roberts et al., 2017; Valavi et al., 2019). Notably, GWR provides a sharp counterexample: despite competitive performance under random cross-validation, its accuracy degrades substantially under spatial blocking (Fig. 2D), indicating reliance on geographically local interpolation that does not transfer when nearby neighbors are withheld. In contrast, ZeGNN retains high performance under blocked folds (Fig. 2D), consistent with learning regime-conditioned mechanisms that generalize across space. Beyond accuracy, the residual diagnostics indicate that the model absorbs spatial autocorrelation that would otherwise persist as systematic error. Residual Moran's I (reported with residual maps) is substantially attenuated for the ZeGNN compared to baselines (Fig. 2G), indicating that the regime-switching thermodynamic representation internalizes spatial structure rather than post-hoc smoothing it (Moran, 1950; Anselin, 1995).

The thermodynamic diagnostics (global feature importance, phase space, role heterogeneity, and local sensitivity distributions; Fig. 2E) and spatial sensitivity maps (Fig. 2F) show that the recovered drivers align with the synthetic mechanism by which covariates perturb the latent state. At the same time, the regime maps and entropy-based gating uncertainty behave as expected for probabilistic mixtures: regime cores appear as near-deterministic assignments, whereas uncertainty peaks along transition interfaces where multiple regime explanations receive comparable support (Shannon, 1948; Bishop, 2006). These results establish that the ZeGNN can recover interpretable thermodynamic components, represent sharp regime-dependent heterogeneity, and demonstrate superior spatial transfer under blocked cross-validation relative to GWR (Fig. 2D), thereby motivating the subsequent cross-domain applications in which the true decomposition is unknown, but the same diagnostics can be used to interrogate heterogeneous spatial mechanisms. The complete set of synthetic datasets used for validation, including the Global Linear and Nonlinear variants beyond the primary example summarized in Fig. 2, together with their regime construction, covariate generation, and closed form definitions of E and S, are documented in Supplementary Note 1.

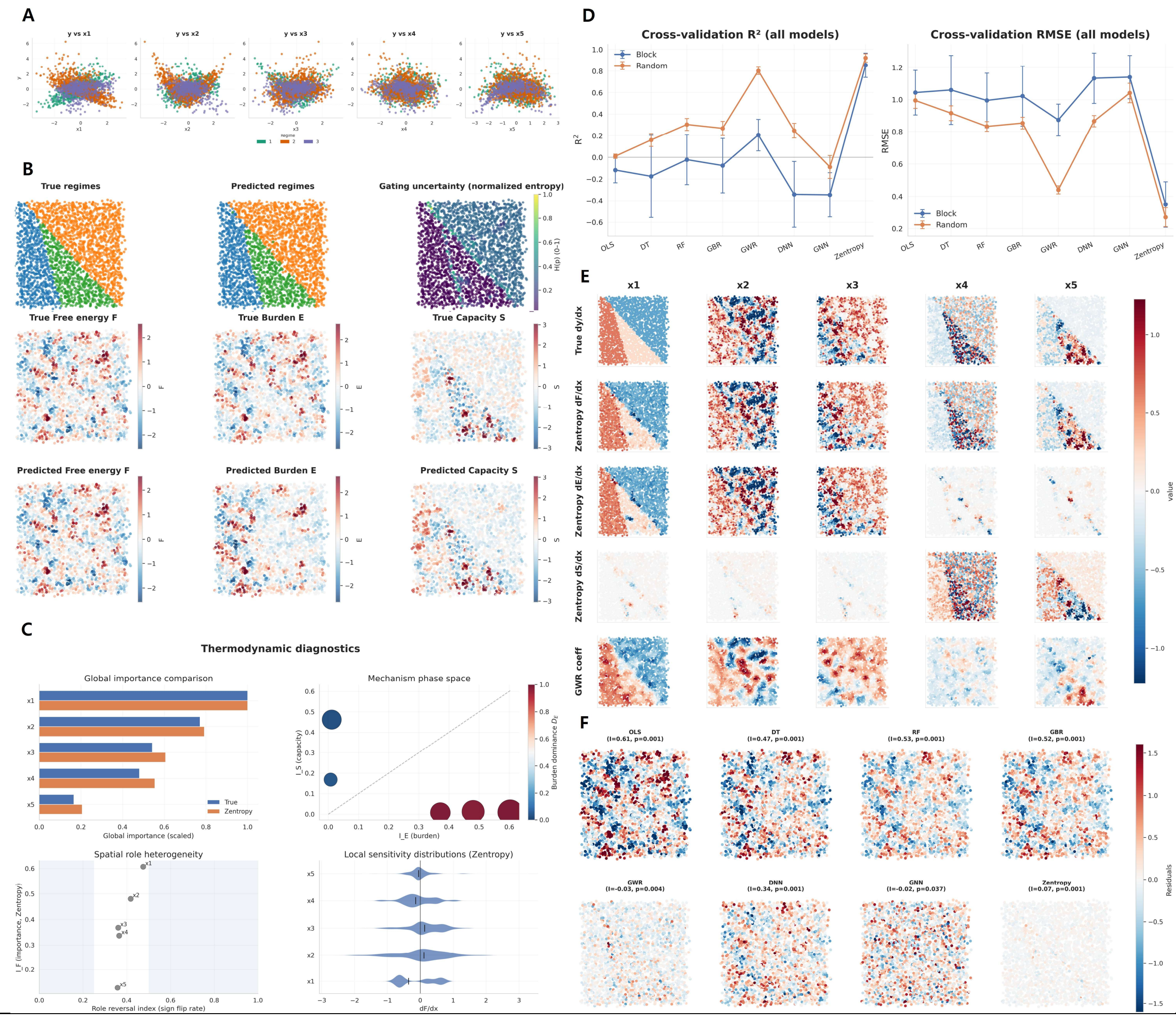

**Fig. 2. Synthetic validation of thermodynamic decomposition, regime identification, and spatial generalization under spatially blocked cross-validation.**
(A) Regime-stratified scatterplots (y vs. x1–x5) illustrating regime-dependent nonlinear response forms in the synthetic data-generating process. (B) Regime recovery and uncertainty diagnostics from ZeGNN: true regimes, predicted regimes, and gating uncertainty (normalized Shannon entropy), alongside standardized maps of the true and predicted outcome y and the predicted thermodynamic fields (free energy $F$, burden $E$, capacity $S$). (C) Thermodynamic diagnostics summarizing global feature importance, mechanism phase space (burden dominance), spatial role heterogeneity, and local sensitivity distributions. (D) Predictive performance across models under random and spatially blocked 5-fold cross-validation ($R^2$ and RMSE). Crucially, GWR shows a pronounced performance drop under spatial blocking, whereas ZeGNN preserves predictive skill, indicating spatial transfer beyond local kernel-based interpolation. (E) Spatial sensitivity maps by covariate (x1–x5), comparing the true local gradient of F and the learned gradient-based sensitivities from ZeGNN (for $F$, $E$, and $S$), with GWR coefficient surfaces shown for reference. (F) Representative residual maps (with Moran's I and p-values) across models, highlighting how explicitly modeling spatial dependence and regime structure reduces residual spatial autocorrelation. Uncertainty is the normalized Shannon entropy of regime probabilities and ranges from 0 to 1.

## Cross-domain predictive performance

We benchmarked ZeGNN against seven baselines spanning global linear regression (OLS), local spatial smoothing (GWR), tree-based learners (DT, RF, GBR), and deep learning models (DNN, GNN) across three domains (Housing, Mental Health, PM2.5) (Fig. 3). To make the evaluation reflect geographic transfer, we report both random 5-fold cross-validation and spatially blocked 5-fold cross-validation: random folds can be overly optimistic when nearby locations appear in both train and test sets, whereas spatial blocking reduces this leakage and better approximates prediction to new places (Roberts et al., 2017; Valavi et al., 2019). Consistent with this rationale, performance systematically degrades under spatial blocking (Perf. Gap), revealing how strongly each model relies on short-range spatial dependence. As shown in Fig. 3 and reported numerically in Supplementary Table S1, ZeGNN achieves the strongest spatial cross-validation performance across all three case studies, with the highest spatially blocked $R^2$ and the lowest spatially blocked RMSE in Housing, Mental Health, and PM2.5. By contrast, some models that perform strongly in-sample or under random cross-validation, particularly GWR, show substantially weaker transfer under spatial blocking in some cases, underscoring the importance of evaluating spatial generalization directly.

This contrast is intuitive especially for GWR because it is explicitly calibrated with geographically local kernels. Under random cross-validation, each test unit typically retains very close training neighbors and yields high apparent accuracy. Under spatial blocking, those neighbors are withheld and performance can collapse. This effect is most pronounced in the Mental Health case, where GWR drops from random cross-validation $R^2$ of 0.906 to spatial cross-validation $R^2$ of 0.065. This highlights its sensitivity to neighbor availability and spatial domain shift. In contrast, ZeGNN shows substantially smaller degradation while retaining strong spatial transfer in the two domains (Housing and Mental Health). In Housing, the spatial cross-validation $R^2$ is 0.838. In Mental Health, the spatial cross-validation $R^2$ is 0.770. This pattern is consistent with its goal of representing spatial dependence and regime wise heterogeneity within the model rather than leaving it as residual structure.

The PM2.5 case presents a fundamentally harder transfer setting. Several models achieve high random cross-validation accuracy, but many show sharp drops in spatial block cross-validation. This pattern is consistent with the physics of air pollution fields. Monthly mean PM2.5 is dominated by large-scale transport and meteorology-driven gradients that are spatially coherent. Random splits inadvertently exploit strong local similarity, whereas spatial blocks remove that context and force predictions into distinct synoptic and microclimatic regimes. In this regime shifted setting, flexible nonlinear learners such as RF retain comparatively better robustness. Models that depend more directly on neighborhood support, including local smoothing and graph-structured propagation, can degrade more when withheld blocks reduce effective neighborhood information and learned regime structure does not transfer cleanly across space.

Finally, residual Moran's I complements predictive metrics by diagnosing whether models leave systematic spatial structure unexplained. Across domains, global and tree-based models tend to leave stronger positive residual autocorrelation, whereas spatially structured learners, including GWR and graph-based models, attenuate it substantially (Fig. 3). This indicates improved control of spatial dependence rather than simply redistributing it into residuals.

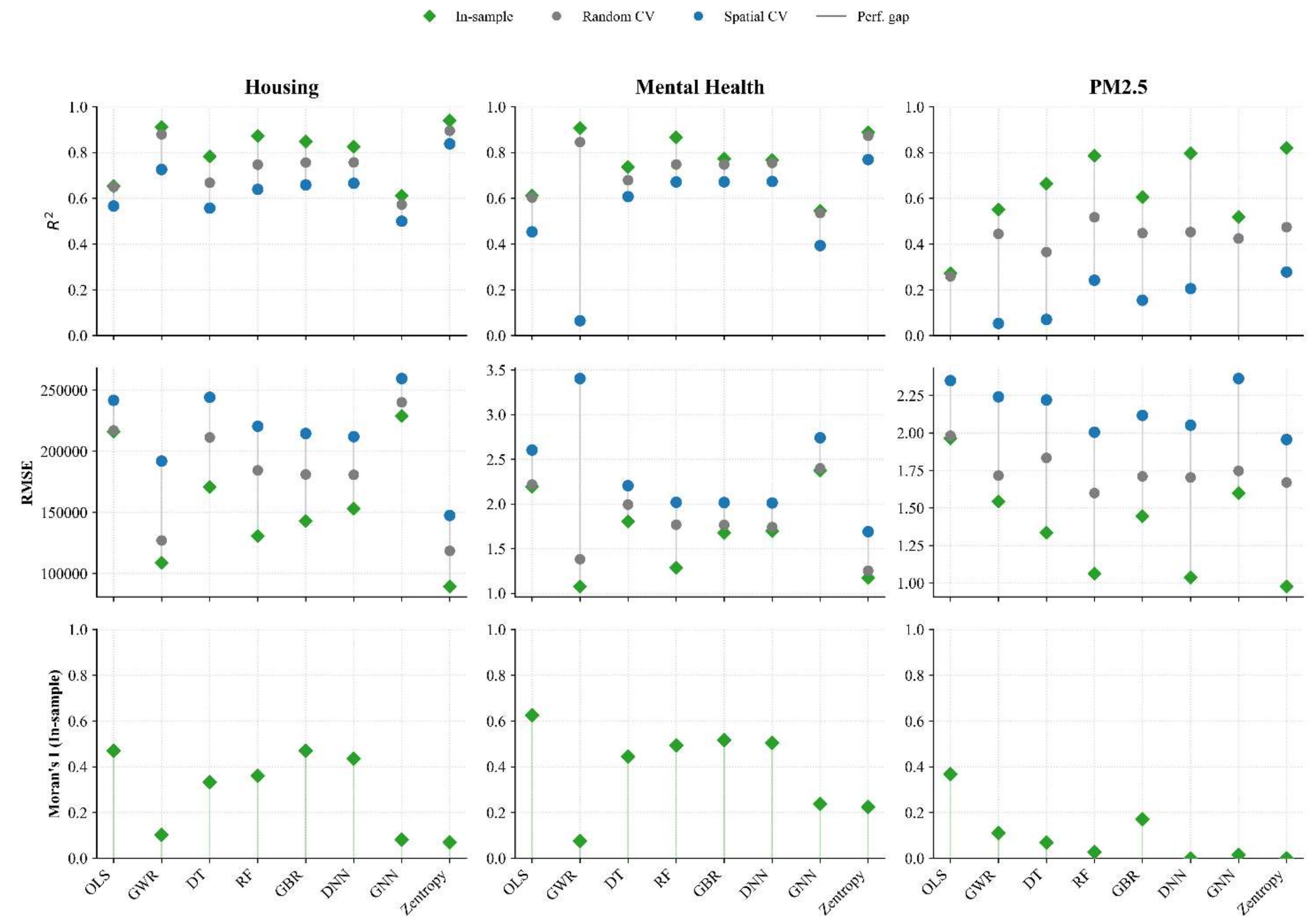

**Fig. 3. Predictive performance, generalization gap, and residual spatial autocorrelation across three domains.** Across Housing, Mental Health, and PM2.5, we summarize performance for eight models (OLS, GWR, DT, RF, GBR, DNN, GNN, ZeGNN) using $R^2$ (top row) and RMSE (middle row) under three evaluation regimes: in-sample (green diamonds), random 5-fold cross-validation (grey circles), and spatial-block 5-fold cross-validation (blue circles). Vertical segments ("Perf. gap") span the in-sample to spatial-block values to visualize degradation under geographic transfer. The bottom row reports Global Moran's I of in-sample residuals. The exact values of each data point in the figure can be found in the Supplemental Table S4.

**Spatial thermodynamic regimes and uncertainty structure**

The ZeGNN framework demonstrates that the high-dimensional heterogeneity of spatial systems is not purely stochastic noise but can be structured into geographically coherent latent regimes, consistent with long-standing views that spatial data often exhibit both spatial dependence and spatial heterogeneity rather than a single global mechanism (Anselin, 1988; LeSage & Pace, 2009). The model partitions study domains into contiguous patches, aligning with the broader idea that "near things are more related than distant things" and that spatial processes can vary by place (Tobler, 1970; Fotheringham et al., 2009). This regime-type organization is also consistent with spatial analysis traditions that explicitly target local pockets of non-stationarity and clustering (Getis & Ord, 1992; Anselin, 1995) and with spatial-econometric approaches that estimate spatial regimes as internally homogeneous but mutually distinct mechanisms (Billé et al., 2017; Andreano et al., 2017; Anselin & Amaral, 2024). Instead of enforcing a single global relationship, the model allows the system to settle into distinct local stability zones where the functional relationship between Burden $E(x)$ and Capacity $S(x)$ remains internally consistent, similar to regime-based interpretations of spatial heterogeneity (Bille et al., 2017; Anselin & Amaral,2024).

Notably, the gating uncertainty, computed from regime probabilities using an entropy functional (Shannon, 1948), provides a spatial diagnostic of ambiguity that naturally arises in probabilistic mixture formulations (Jordan & Jacob, 1994; Bishop, 2006). Because mixture models represent observations as weighted combinations of competing latent explanations, uncertainty is expected to rise where multiple regime explanations receive comparable support. In spatial settings, such ambiguity is especially informative at regime interfaces, which are widely recognized as transition zones where the assumption of a single stable local mechanism is least defensible (Bille et al., 2017; Anselin & Amaral, 2024). This framing also resonates with boundary-sensitive spatial smoothing and segmentation ideas, where structural breaks and edges concentrate modeling difficulty.

In the Housing data analysis (Fig. 4A), this thermodynamic organization manifests as localized clustering that tessellates King County into structurally distinct submarkets, conceptually consistent with the spatial econometric view that housing processes can be governed by distinct regime logics across space (Billé et al., 2017; Anselin & Amaral, 2024). The decomposition provides a mechanistic basis for valuation beyond price prediction. As shown in the regime-stratified scatterplots (Fig. 4A, right panels), the relationships between the observed outcome and the learned free energy, burden, and capacity differ across regimes, and the regime stratified distributions of these quantities are also distinct, consistent with the interpretation of regimes as qualitatively different generating mechanisms rather than a smooth continuum (Andreano et al., 2017; Fotheringham et al., 2009). The sharp separation in the $E-S$ plane supports that these regimes represent fundamental shift in market logic, with uncertainty peaking precisely at friction zones where competing valuation mechanisms interact, which is an expected pattern when probabilistic mixtures encode ambiguity (Jordan & Jacobs, 1994; Bishop, 2006).

This thermodynamic consistency extends to the macroscopic regional organization observed in the mental health domain (Fig. 4B). Unlike localized housing submarkets, mental health regimes from massive, contiguous blocks that span across state lines, suggesting that structural drivers of social vulnerability can operate at broad spatial scales (Anselin, 1988; LeSage & Pace, 2009). Within these broad clusters, the regime-wise distributions follow a burden–capacity logic consistent with the stress-buffering hypothesis, which posits that resources can attenuate the adverse effects of stressors (Cohen & Wills, 1985). The concentration of uncertainty at boundaries of these large-scale zones is consistent with regime-mixing and gradual spatial gradients rather than disjointed local effects or random outliers, a core motivation for spatially varying relationship models (Brunsdon et al., 1996; Fotheringham et al., 2009).

The regime structure also captures broad-scale coherence even under sparse monitoring geometry (Fig. 4C). In the PM2.5 case, the inferred regimes exhibit regional organization consistent with the expectation that environmental processes vary across space and that local relationships can differ by context (Brunsdon et al., 1996; Fotheringham et al., 2009). Regions of elevated free-energy-like pressure $F(x)$ align with areas where burden overwhelms capacity, and the spatial contiguity of regimes suggests the model is detecting structured spatial heterogeneity rather than pointwise overfitting. By localizing heterogeneity into interpretable pathways, ZeGNN reframes uncertainty from a residual artifact into a structural feature, providing a diagnostic of where mechanism attribution is spatially stable versus where the system resides in a transitional state, which is precisely the interpretive value of probabilistic regime assignments (Shannon, 1948; Jordan & Jacobs, 1994; Anselin & Amaral, 2024).

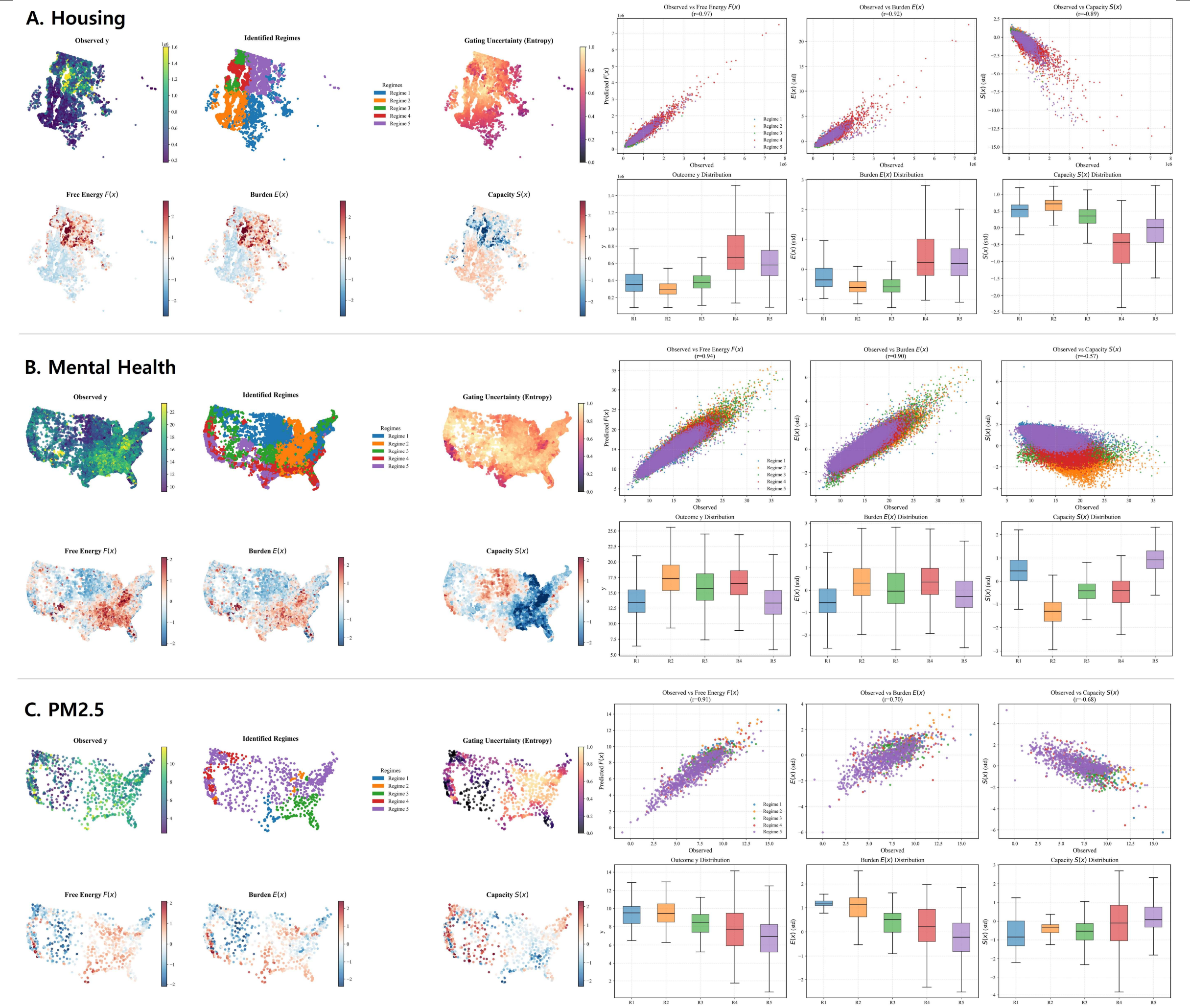

**Fig. 4 | Spatial thermodynamic regimes, gating uncertainty, and burden-capacity decomposition across domains**

(A) Housing (King County), (B) Mental health (tract-level), and (C) PM2.5 (June 2023) applications. For each domain, the left panels summarize the identified latent regimes (hard assignment by $\text{argmax}_k\, p_k(x)$) and gating uncertainty (normalized Shannon entropy of $p_k(x)$), alongside standardized maps of the learned free energy $F(x)$, burden $E(x)$, and capacity $S(x)$. The right panels report domain-wise diagnostic relationships between the observed outcome and the learned thermodynamic components, including observed vs. $F(x)$, $E(x)$, and $S(x)$, and regime-stratified distributions of the outcome and decomposition terms. Regime labels are nominal and do not imply ordering. $F$, $E$, and $S$ are shown in standardized units for within domain spatial comparison. Uncertainty ranges from 0 to 1.

To make these transition dynamics explicit, we visualize the full regime probability surfaces produced by the gating network, $P(Regime\ k\ |x)$ (Fig. 5). These probability maps provide a direct interpretation of the entropy-based uncertainty diagnostic reported in Fig. 4. Regime cores appear as spatial zones with near-deterministic assignment, where one regime dominates with high probability, whereas interfaces are characterized by spatially coherent mixing in which two or more regimes attain comparable probabilities. This behavior is expected in mixture-based classification setting, where uncertainty naturally peaks near decision boundaries and mixed-probability regions (Bezdek, 1981; Jordan & Jacobs, 1994; Bishop, 2006). Importantly, these transition zones are not scattered point anomalies. Instead, they form geographically structured corridors consistent with gradual shifts in latent state, echoing classic spatial continuity arguments (Tobler, 1970) and regime-based representations in which heterogeneity is expressed through organized, interpretable spatial partitions (Anselin & Amaral, 2024).

Across the three domains, the $P(Regime\ k\ |x)$ fields reinforce this interpretation at different spatial scales. In the Housing case, probability mass concentrates into compact submarket cores with narrow mixing belts, consistent with localized regime tessellation. In the mental health case, probability gradients span broad regions and intensify at interstate-scale boundaries, consistent with large-scale socioeconomic transitions. In the PM2.5 case, probability fields organize into synoptic-scale structures, indicating that regimes reflect regional meteorological control rather than pointwise irregularity. These probability maps complement hard regime assignments by revealing how strongly each location belongs to a regime, and therefore why entropy peaks exactly where the model's explanation is intrinsically ambiguous.

This interpretation is also supported by the fitted regime-usage diagnostics reported in Supplementary Table S5. Housing retains all five regimes at the global level ($n_{eff,global}$ = 5.0), with a relatively high average local effective regime count ($n_{eff,local}$ = 3.425) and a modest maximum dominant-regime share (0.273), consistent with broad multi-regime mixing. Mental Health shows a similarly distributed regime structure ($n_{eff,global}$ = 4.986, $n_{eff,local}$ = 3.611, max dominant-regime share = 0.318). By contrast, PM2.5 exhibits fewer effectively used regimes ($n_{eff,global}$ = 3.867), a lower local effective regime count ($n_{eff,local}$ = 2.687), and a substantially larger dominant-regime share (0.673), consistent with the sharper regime cores and more uneven spatial partitioning visible in Fig. 5.

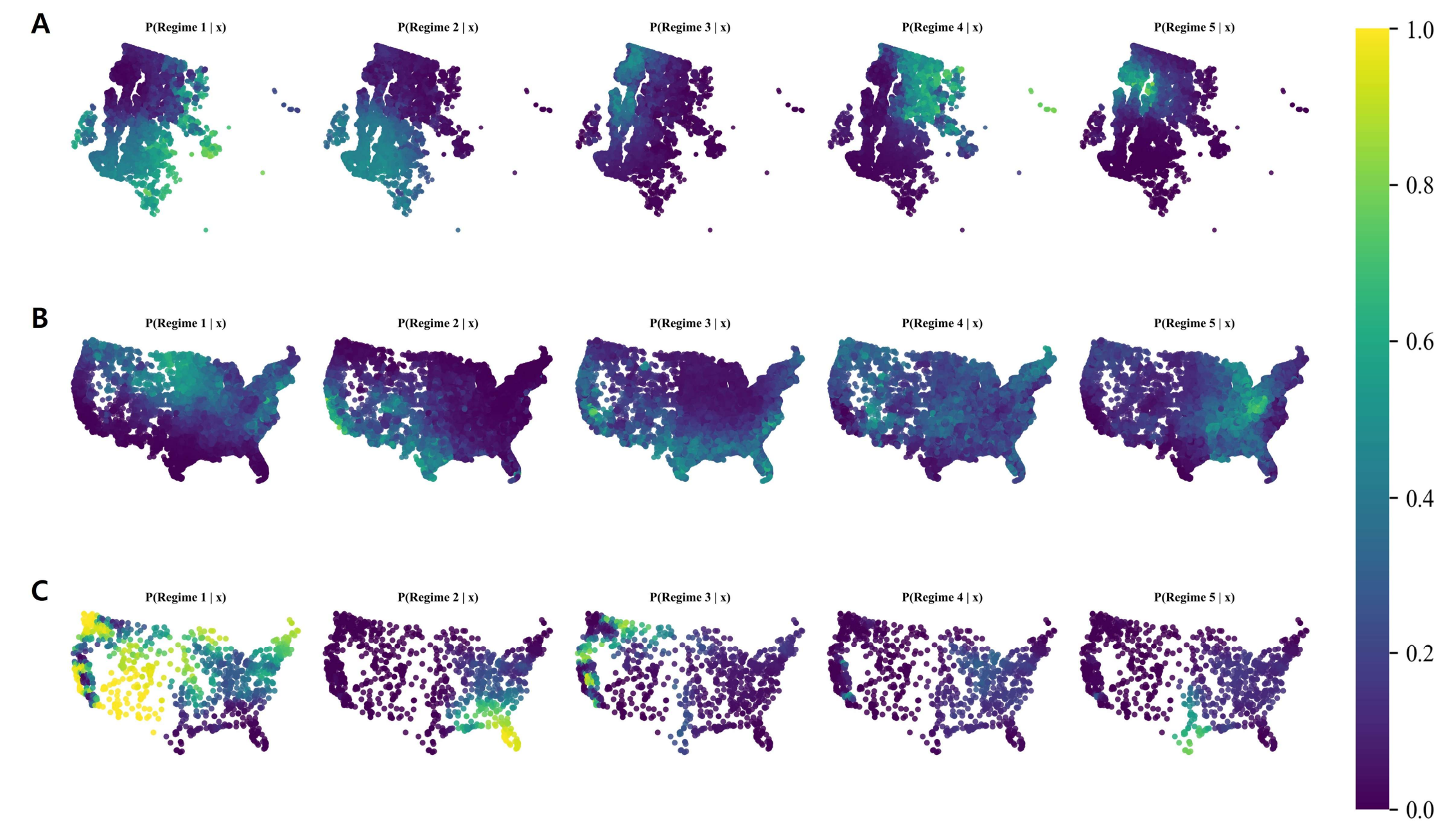

**Fig. 5 | Regime probability maps reveal soft thermodynamic partitions and transition corridors.**
(A) Housing ($K_{upper}$=5), (B) Mental Health ($K_{upper}$=5), (C) PM2.5 ($K_{upper}$=5). For each case, panels show the gating network outputs $P(Regime\ k \mid x)$ evaluated at observation locations. Warmer colors indicate higher probability. Regime cores correspond to spatial zones where one regime dominates (high $P$), while transition zones appear where multiple regimes exhibit moderate probabilities simultaneously, consistent with peaks in entropy-based gating uncertainty (Fig. 4). The spatial coherence of mixed zones confirms that heterogeneity is organized into structured thermodynamic regimes rather than pointwise noise.

**Thermodynamic marginal effect maps expose regime-dependent pathways and role reversals**

Fig. 6 extends standard variable-importance summaries into mechanistic statements about how covariates perturb the learned thermodynamic state. It achieves this by jointly summarizing pathway dominance, spatial stability, and representative decomposition maps of local sensitivities ($\partial F/\partial x$, $\partial E/\partial x, \partial S/\partial x$). Across all three domains, the diagnostics reveal a recurring empirical structure: variables that dominate the net thermodynamic response are typically those whose local perturbations are not strongly offset by an opposing adjustment in other component. In contrast, variables with pronounced regime-dependence tend to exhibit spatially organized compensation patterns where burden and capacity responses co-occur and partially counterbalance.

In the Housing system, the strongest mechanistic split is anchored by Grade versus size-related attributes like sqft living and sqft above (Fig. 6A). Grade serves as a capacity-dominant mechanism in the phase space and plays a comparatively spatially consistent role. Conversely, sqft living and related size measures cluster as burden-dominant drivers. The scientific consequence lies in the pathway signature rather than the ranking alone. The sqft_living maps show broad regions where the net sensitivity aligns with a burden-side response, while the net response for Grade is expressed through a distinct capacity-side pattern. This produces a spatially interpretable separation between quantity-driven and quality-driven valuation pathways. Moreover, neither variable is perfectly pathway pure. Each exhibits non-trivial spatial structure in the other component, which is most visible in geographically coherent patches rather than random noise. This phenomenon represents the empirical fingerprint of local compensation, where changes that would otherwise amplify the net state are partially offset by a countermovement in the opposing component.

These cross-component patterns are both expected and interpretable, even though covariates were pre-classified into burden-like and capacity-like blocks. The pre-classification serves as an inductive bias about direct routing rather than a hard constraint that forbids cross-component responses. The model design explicitly allows variable roles to be diagnosed post hoc from learned sensitivities instead of assuming they remain fixed. Fig. 6 clarifies that cross-channel responses concentrate precisely where the regime explanation is least settled. These responses appear along spatially coherent transition corridors where multiple regimes compete rather than within regime cores. This behavior aligns with the regime-probability patterns observed in Fig. 5, where uncertainty rises at interfaces because mixture explanations overlap.

In practical terms, a variable assigned to the burden block can still shift the effective capacity surface by reweighting which regime's capacity structure dominates locally. A capacity-classified variable can similarly reshape the apparent burden field through this same regime competition mechanism. This appears in Fig. 6 as structured non-zero sensitivities in both components alongside elevated role heterogeneity for the same variables. This pattern is especially visible for role-heterogeneous examples (e.g., sqft_lot in Housing). Such evidence suggests that the underlying mechanism varies most where the system is near a regime boundary.

The same discovery generalizes beyond Housing. In the Mental Health case, Neighborhood Disadvantage emerges as a dominant burden-like driver with a nation-scale positive net response (Fig. 6B). Yet its maps simultaneously show an organized opposing signal in the capacity component. This is consistent with a compensatory shift that is strongest in regions exhibiting regime-dependence rather than uniformity. NatureParks and Casinos provide representative capacity-dominant and role-heterogeneous pathways, respectively. In the PM2.5 case, mean sea level pressure (MSLP) concentrates as a burden-dominant driver, while boundary layer height (BLH) appears as a capacity-dominant stabilizer and near-surface zonal wind (U10) exhibits pronounced role heterogeneity (Fig. 6C).

However, the June 2023 wildfire event serves as a critical stress test for distinguishing mechanistic reasoning from mere statistical interpolation. While baseline models like Random Forest (RF) achieve competitive predictive accuracy by effectively memorizing these extreme anomalies, ZeGNN explicitly diagnoses this period as a phase transition into a burden-dominated regime. Specifically, the sensitivity analysis (Fig. 6C and Supplementary Fig. S10) reveals a regime-specific surge in the contribution of thermodynamic stressors to the Burden term ($E$), effectively suppressing the stabilizing influence of the Capacity term ($S$). This suggests that the model is capturing a structured regime shift associated with the anomaly, rather than merely reproducing elevated PM2.5 concentrations. Importantly, both exhibit spatially structured cross-component responses that intensify in regime-dependent zones. This capability to disentangle the source of the anomaly (burden surge) from the state of the system (high PM2.5) demonstrates that the framework operates as a diagnostic instrument rather than a mere curve-fitting tool.

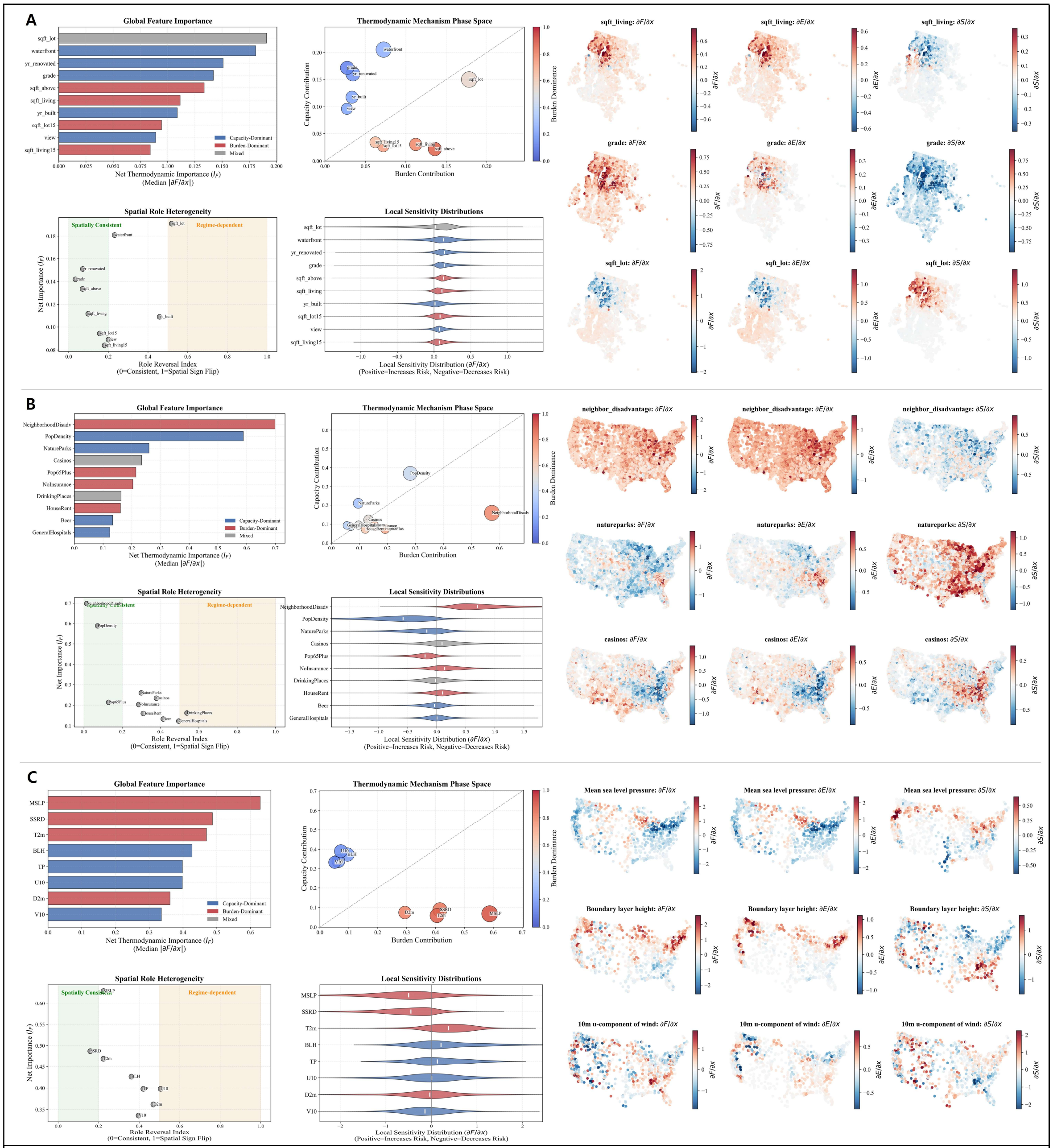

**Fig. 6 | Thermodynamic sensitivity diagnostics and representative decomposition maps**

(A) Housing, (B) Mental Health, and (C) PM2.5. For each case, the left 2×2 diagnostic panels summarize: Global Feature Importance (median $|\partial F/\partial x|$), Thermodynamic Mechanism Phase Space ($I_E$ vs. $I_S$ with size $\propto I_F$ and color indicating burden dominance $D_E$), Spatial Role Heterogeneity (Role Reversal Index vs. $I_F$), and Local Sensitivity Distributions of $\partial F/\partial x$. The right-side maps visualize $\partial F/\partial x$, $\partial E/\partial x$, and $\partial S/\partial x$ for three representative variable selected to highlight a burden-dominant, capacity-dominant, and high role-reversal mechanism: (A) sqft_living, grade, and sqft_lot; (B) Neighbor_Disadvantage, NatureParks, and Casinos; (C) MSLP (mean sea level pressure), BLH (boundary layer height), and U10 (10 m u-component of wind). The full decomposition map is provided in Supplementary Fig S6-8.

## Discussion

This study demonstrates that a thermodynamic structural prior offers a practical route to interpretable

GeoAI under spatial heterogeneity. Rather than attaching explanations to a black-box predictor, ZeGNN embeds interpretability within the predictive structure itself by expressing prediction as a balance between two latent but meaningful functions. This design aligns with arguments that high-stakes scientific and policy settings require interpretability built into the model itself rather than relying on post-hoc rationales (Rudin, 2019).

A central implication of ZeGNN is that interpretation becomes a property of the model state variables rather than an afterthought. While post-hoc attribution methods like local surrogate explanations are widely used (Ribeiro et al., 2016; Lundberg & Lee, 2017), they can fail silently when explanations are not faithful or are sensitive to irrelevant perturbations. Research on sanity checks has formalized this concern by showing that certain saliency-style explanations can remain visually similar even after model parameter randomization (Adebayo et al., 2018). In contrast, our burden–capacity decomposition produces interpretable intermediate fields ($E$ and $S$) whose spatial structure is directly tied to the predictive representation. This enables a mechanistic reading of why a location is burden-dominant, capacity-dominant, or uncertain.

Methodologically, the framework sits between purely statistical spatial modeling and fully physics-constrained learning. Physics-informed machine learning typically encodes governing equations or conservation laws to restrict function classes (Raissi et al., 2019; Karniadakis et al., 2021). However, the constraint is weaker but more broadly applicable as the predicted state is organized around a free-energy relation that explicitly represents opposing pathways. This soft-physics inductive bias is especially valuable in geographic systems where the governing dynamics are only partially known or vary across space in ways that frustrate global parametric assumptions.

The regime structure and its uncertainty also carry an interpretation beyond clustering. The gating probabilities form a continuous surface where uncertainty concentrates in coherent transition belts rather than isolated points. This spatial organization resonates with work on critical transitions where heightened variability and reduced stability often occur near boundaries between alternative states (Scheffer et al., 2009; Dakos et al., 2012; Kéfi et al., 2014). Although our analysis is not a time-evolving early-warning study, the maps provide an analogous spatial diagnostic. Where multiple regimes compete, mechanistic attribution becomes less stable and local sensitivities broaden. This suggests locations where small covariate changes may be most consequential for shifting the thermodynamic balance.

From a geospatial modeling perspective, these results speak to recurring obstacles regarding nonstationarity and limited transportability across contexts. Recent syntheses emphasize that geospatial machine learning must contend with spatial dependence and heterogeneity simultaneously while maintaining interpretability for environmental research (Jemeļjanova et al., 2024; Koldasbayeva et al., 2024). However, in the era of increasing climate volatility, which is exemplified by the 2023 wildfire crisis, predictive accuracy alone is insufficient. The core contribution of the ZeGNN framework lies in its ability to rationalize spatial heterogeneity through a thermodynamic narrative, elevating GeoAI from a passive predictive instrument to an abductive scientific tool. By combining graph-based contextualization with regime-specific thermodynamic structure, ZeGNN provides a concrete pathway for reconciling predictive accuracy with geographically explicit and mechanism-oriented narratives that remain consistent across domains.

Several limitations deserve emphasis. The decomposition depends on how covariates are assigned to burden versus capacity channels. While we used domain knowledge to motivate these choices, our results (Fig. 6) demonstrate that the model is robust to this initialization. The emergence of role heterogeneity confirms that the ZeGNN recovers the effective functional role from the data itself, independent of the initial label, thereby guiding the solution toward interpretability without sacrificing data-driven flexibility. Additionally, the local gradients quantify associations in the learned model rather than causal effects and should be interpreted as mechanistic hypotheses. Furthermore, evaluation remains sensitive to how folds are constructed despite the use of spatially blocked cross-validation (Roberts et al., 2017; Valavi et al., 2019). Relatedly, information criteria such as AIC are most natural for likelihood-based models and should be interpreted cautiously when juxtaposed with flexible neural learners (Burnham & Anderson, 2002).

Despite these caveats, ZeGNN opens clear paths forward. Extending the model to spatiotemporal graphs would allow regimes and thermodynamic sensitivities to evolve over time. Another direction is to incorporate stronger physical or institutional constraints when available to combine the generality of the thermodynamic prior with domain-specific structure. Finally, the thermodynamic perspective used here is part of a broader effort to incorporate energy-based structure into data-driven learning. Zentropy theory was developed to predict thermodynamic free energy without fitting parameters (Liu, 2024), and recent work proposes thermodynamics-inspired neural frameworks for heterogeneous data-driven modeling (Wang et al., 2026). Together with our findings, this trajectory suggests that thermodynamic decompositions can function as a general-purpose mechanism scaffold for heterogeneous geospatial systems. This supports interpretation not by simplifying the model but by organizing complexity into scientifically legible components.

## Methods

Since the conceptual architecture and interpretive role of ZeGNN are introduced in the result section and summarized in Fig. 1, this section focuses on the implementation details needed for reproducibility. Specifically, we describe outcome standardization, spatial graph construction, neural parameterization and optimization, hyperparameter selection under five-fold spatially blocked cross-validation, post-training thermodynamic diagnostics, and the baseline models used for comparison.

All models, including baselines and ZeGNN, map standardized feature $x_i$, to a standardized outcome $z_i$. Let $y_i$ denote the observed outcome and let $\mu_y$ and $\sigma_y$ be the mean and standard deviation of $y$ in the training set. We define the standardized outcome as:

$$z_i = \frac{y_i - \mu_y}{\sigma_y} \tag{6}$$

Models are trained to predict $\hat{z}_i$, and predictions are transformed back to the original scale by $\hat{y}_i = \mu_y + \sigma_y \widehat{z_i}$ for reporting and evaluation.

### Spatial graph construction

We represent the study domain as a graph $G = (V, E)$, where each $i \in V$ corresponds to a spatial unit and edges $E$ connect nearby (or otherwise strongly linked) units. To encode local spatial context, we construct a k-nearest neighbor (kNN) graph on the point coordinates. Let $c_i \in \mathbb{R}^2$ denote the projected planar coordinates of unit $i$. For each $i$, we identify its k nearest neighbors by Euclidean distance in $(c_x, c_y)$ space.

Because the neighborhood size $k$ defines the model's spatial inductive bias, we selected graph topology using 5-fold cross validation rather than treating $k$ as arbitrary (Stakhovych & Bijmolt, 2009; Roberts et al., 2017; Valavi et al., 2019). For each candidate graph, we record held-out blocks ($R^2_{CV}$) and residual spatial autocorrelation on held-out predictions (Moran's I), and we selected the final graph using 1-SE parsimony rule that favors the smallest admissible $k$ with residual spatial autocorrelation treated as a secondary diagnostic (Moran, 1950; Hastie et al., Importantly, $k$ controls only the graph topology (spatial neighborhood size) and is conceptually distinct from the number of latent regimes upper bound $K$ used in the ZeGNN search.

This yields an undirected adjacency matrix $A \in \mathbb{R}^{N \times N}$ with entries:

$$A_{ij} = \begin{cases} 1, & if\ j \in N_k(i)\ or\ i \in N_k(j), \\ 0, & otherwise, \end{cases} \tag{7}$$

Where $N_k(i)$ is the neighbor set of node $i$. We then construct a row-normalized (random-walk) adjacency used for mean-aggregation message passing:

$$\tilde{A} = D^{-1}A \tag{8}$$

Where $D$ is the diagonal degree matrix with $D_{ii} = \sum_j A_{ij}$. This choice matches the implementation's mean neighborhood aggregation. In implementation, $\tilde{A}$ serves as the diffusion operator used to serves as the diffusion operator used to inject neighbourhood context into the gating network.

**Implemented neural parameterization**

As introduced in the Results and summarized in Fig. 1, ZeGNN combines graph-aware soft regime gating with thermodynamic burden and capacity channels. Here we retain only the implementation details needed for reproducibility. For a given regime upper bound $K_{upper}$, the model returns the final free-energy prediction together with the mixture-averaged burden field, the mixture-averaged capacity field, the regime probabilities, and the normalized gating entropy. $K_{upper}$ denotes the upper bound on the number of latent regimes available to the model, rather than a direct estimate of the true number of regimes in the data. For each spatial unit $i$, the gating network produces a probability vector $(\pi_{i1}, \ldots, \pi_{iK})$ over these candidate regimes, where $\pi_{ir}$ is the soft membership of unit $i$ in regime $r$, $\pi_{ir} \geq 0$, and $\sum_{r=1}^{K_{upper}} \pi_{ir} = 1$. In this sense, $K_{upper}$ is a model-capacity parameter that specifies the maximum number of regimes the mixture may use. After fitting, regime usage is summarized diagnostically rather than treated as a stand-alone selection rule. The global effective regime count summarizes how many regimes are effectively used across the study domain, whereas the local effective regime count summarizes the average degree of regime mixing within spatial units. Accordingly, these effective counts may be smaller than $K_{upper}$ when some candidate regimes receive negligible occupancy or when individual locations are dominated by only a subset of the available regimes.

Let $x_i^E \in \mathbb{R}^{pE}$ and $x_i^S \in \mathbb{R}^{pS}$ denote the standardized burden and capacity feature blocks at location $i$. In the implemented model, burden and capacity are parameterized through separate neural channels rather than through a single shared predictor. The burden channel uses a one-hidden-layer encoder followed by a linear regime head to produce $K_{upper}$ burden outputs, and the capacity channel uses the same structure to produce $K_{upper}$ capacity outputs. Thus, the model produces regime-indexed latent outputs $E_{ik}$ and $S_{ik}$ for each location.

A regime-specific temperature vector is learned jointly with the burden and capacity channels. In implementation, this parameter is stored in unconstrained form and transformed to the positive domain before entering the forward pass, ensuring that the regime temperatures remain strictly positive. The regime-specific free-energy terms and the entropy-adjusted final prediction follow the formulation introduced in the Results.

The regime probabilities are generated by a graph-aware gating network. For each spatial unit, we concatenate the standardized predictors and standardized coordinates, and pass this augmented input through a multilayer perceptron with two hidden layers. The resulting regime logits are then diffused over the row-normalized spatial graph using mean aggregation, and a softmax transformation is applied to obtain the regime probabilities $p_{ik}$. This implementation allows neighboring units to share contextual information before regime assignment, while preserving location-specific variation in the final gating surface.

These components define a graph-aware mixture-of-regimes model in which burden, capacity, temperature, and regime probability are estimated jointly. The forward pass returns the entropy-adjusted free-energy prediction $F_i$, the mixture-averaged burden and capacity fields $E_i$ and $S_i$, the regime probability surface $p_{ik}$, the regime-specific latent outputs, and the normalized node-level entropy used for post-training diagnostics.

**Hyperparameter selection**

Hyperparameter selection is performed using 5-fold spatial block cross-validation rather than an elbow-based search over a fixed latent-regime count. In the implemented workflow, model complexity is controlled through the graph neighborhood size $k$, the regime upper bound $K_{upper}$, and the regularization weights ($\lambda_{sparse}, \lambda_{mag}$). Candidate grids are scenario-specific in the simulation experiments. The regularization grid uses $\lambda_{sparse} \in \{0, 0.001, 0.005\}$ and $\lambda_{mag} \in \{0.001, 0.01\}$, with a reduced $\lambda_{mag}$ grid in the Global Linear implementation.

For each candidate configuration, we record held-out spatial $R^2$, RMSE, and residual Moran's I. The final configuration is selected using a 1-SE parsimony rule based on mean spatial $R^2$, favoring the smallest admissible graph neighborhood and the smallest total regularization among models within one standard error of the best-performing configuration. Residual Moran's I and fitted regime-usage diagnostics are retained as secondary diagnostics rather than primary optimization targets. Accordingly, $K_{upper}$ is treated as a capacity parameter in the search, rather than as an elbow-selected estimate of the true number of latent regimes. In the empirical applications, $K_{upper} = 5$ was fixed during the final hyperparameter search, and model selection proceeded over graph neighborhood size and regularization.

**Regime diagnostics**

After fitting, we summarize regime structure using secondary diagnostics rather than stand-alone selection rules. These include the spatial coherence of the fitted regime surface, the feature-space separation of soft or dominant regime assignments, and residual Moran's I from held-out predictions. In the current workflow, regime maps and normalized gating entropy are interpreted jointly, such that core areas correspond to high-probability assignments and transition zones correspond to elevated mixture uncertainty. These diagnostics are reported alongside predictive performance to characterize how the fitted regime structure behaves spatially, but they do not replace blocked cross-validation as the primary basis for model selection.

**Post-training thermodynamic diagnostics**

The implemented model does not impose strict orthogonality between burden and capacity during training. Instead, thermodynamic decomposition is evaluated after fitting through the joint behavior of the learned $F$, $E$, and $S$ surfaces, the regime-conditioned latent outputs, and the derivative-based diagnostics described below. This design allows burden and capacity to remain complementary but not artificially constrained, while still supporting post hoc interpretation of sign structure, local sensitivities, and regime-level thermodynamic patterns.

**Training objective and regularization**

ZeGNN is trained end-to-end with a composite objective consisting of prediction error, an anti-collapse occupancy barrier, and a magnitude penalty on regime-specific latent outputs. Let $\mathcal{T}$ denote the set of training indices. The primary term is a mean-squared error between observed and predicted standardized outcomes:

$$\mathcal{L}_{MSE} = \frac{1}{|\mathcal{T}|} \sum_{i \in \mathcal{T}} (\hat{z}_i - z_i)^2 \tag{9}$$

To discourage regime collapse, we regularize the average regime occupancy rather than directly optimizing node-level entropy as a separate loss term. If

$$\bar{p}_k = \frac{1}{|\mathcal{T}|} \sum_{i \in \mathcal{T}} p_{ik} \tag{10}$$

Denotes the average mixing weight for regime $k$, we apply the anti-collapse penalty

$$\mathcal{L}_{sparse} = -\sum_{k=1}^{K} \log(\bar{p}_k + \epsilon) \tag{11}$$

We additionally penalize the magnitude of the regime-specific burden and capacity outputs:

$$\mathcal{L}_{mag} = \frac{1}{|T|} \sum_{k=1}^{K} \sum_{i \in T} \left(E_i^2 + S_i^2\right) \tag{12}$$

which discourages unbounded growth in the latent thermodynamic channels.

The final objective is:

$$\mathcal{L} = \mathcal{L}_{MSE} + \lambda_{sparse}\mathcal{L}_{sparse} + \lambda_{mag}\mathcal{L}_{mag} \tag{13}$$

Node-level entropy is not optimized as a separate regularization term in the final implementation. Instead, normalized gating entropy is retained as a post-training diagnostic and enters the forward prediction through the entropy-adjusted mixture described in the Results. Optimization uses Adam with learning rate 0.005. In the simulation implementation, the hidden width is 64, the maximum number of training epochs is 800, and early stopping uses patience 60. Within each fold, the best model state is restored before held-out evaluation, and the selected configuration is then refit on the full dataset to generate in-sample maps and downstream diagnostics.

We selected the graph neighborhood size $k$, the regime upper bound $K_{upper}$, and the regularization weights ( $\lambda_{sparse}, \lambda_{mag}$) using 5-fold spatial block cross-validation. To keep the main text concise, the full $k$-selection and regularization sweep summaries are reported in Supplementary Note 2 (Supplementary Fig. S5).

**Gradient-based local effect decomposition**

To quantify how each input variable influences the thermodynamic state, we compute spatial sensitivity fields using automatic differentiation. For each covariate $x_j$, we evaluate:

$$\frac{\partial F}{\partial x_j}, \quad \frac{\partial E}{\partial x_j}, \quad \frac{\partial S}{\partial x_j} \tag{14}$$

At all locations by backpropagation through the trained ZeGNN. We refer to these quantities as local derivative-based effects (marginal effects) because the model is nonlinear and the derivatives summarizes the slope of the learned function around each observation. These gradients reveal not only the total effect of each variable on predicted risk ($\partial F/\partial x_j$) but also decompose it into burden-driven ($\partial E/\partial x_j$) versus capacity-driven ($\partial S/\partial x_j$) pathways. Positive sensitivities in $\partial E/\partial x_j$ indicate that increases in the variable amplify structural burden, while positive sensitivities in $\partial S/\partial x_j$ indicate enhancement of adaptive capacity. To validate gradient accuracy, we perform finite-difference perturbation experiments: for each variable, we perturb standardized inputs by $\delta = 0.1$ and compute $\Delta F = F\left(x + \delta e_j\right) - F(x)$, then verify consistency by spatial correlation between $\Delta F$ and $\partial F/\partial x_j$. High correlation confirms that analytical gradients accurately reflect the model's learned sensitivities (Supplemental Note 1.2.1; Supplemental Fig. S3, S4). As an additional interpretability check, we verify that derivative-based maps are not artifacts of model parameterization by applying standard attribution sanity checks (Adebayo et al., 2018).

**Global thermodynamic importance and functional roles**

Following the gradient-based local effect decomposition, we summarize the spatially varying sensitivities into global, interpretable thermodynamic diagnostics. Let:

$$g_{i,j}^{F} = \frac{\partial F(x_i)}{\partial x_{i,j}}, \quad g_{i,j}^{E} = \frac{\partial E(x_i)}{\partial x_{i,j}}, \quad g_{i,j}^{S} = \frac{\partial S(x_i)}{\partial x_{i,j}} \tag{15}$$

Denote the local gradient components evaluated at location $i$. We then define three complementary metrics that quantify (1) net influence on the free-energy surface, (2) pathway dominance between burden and capacity channels, and (3) stability of effects within regime cores.

We define the net importance of predictor $j$ as the average magnitude of its local effect on the free-energy output:

$$I_F(j) = \frac{1}{N}\sum_{i=1}^{N} \left| g_{i,j}^{F} \right| \tag{16}$$

This metric quantifies how strongly a covariate perturbs the system's net thermodynamic state $F(x)$, aggregating over spatial heterogeneity while remaining independent of sign.

To characterize whether a predictor acts primarily through the burden channel $E$ or the capacity channel $S$, we compute channel-specific importances:

$$I_E(j) = \frac{1}{N}\sum_{i=1}^{N} \left| g_{i,j}^{E} \right|, \quad I_S(j) = \frac{1}{N}\sum_{i=1}^{N} \left| g_{i,j}^{S} \right| \tag{17}$$

And define the burden-dominance index

$$D_E(j) = \frac{I_E(j)}{I_E(j) + I_S(j)} \tag{18}$$

Values $D_E(j) \to 1$ indicate burden-dominant drivers (effects primarily transmitted through $E$), whereas $D_E(j) \to 0$ indicate capacity-dominant mechanisms (effects primarily transmitted through $S$). This provides a post-training, data-driven thermodynamic role characterization for each predictor.

Because ZeGNN represents the spatial system as a gated mixture of regimes, regime interfaces correspond to locations where multiple regime explanations are similarly plausible. In these transition corridors, the regime probability vector $p_i$ is diffuse rather than concentrated on a single regime, producing higher Shannon entropy. Sensitivities evaluated at such locations can reflect not only within-regime functional response but also rapid changes in the mixture weights, and therefore may be less structurally stable for mechanism attribution. To emphasize mechanisms that operate consistently within stable regime interiors, we compute an entropy-weighted importance that down-weights high-entropy locations:

$$I_{core}(j) = \frac{\sum_{i=1}^{N} w_i \left| g_{i,j}^{F} \right|}{\sum_{i=1}^{N} w_i}, \qquad w_i = 1 - \widetilde{H}(x_i) \tag{19}$$

Where $\widetilde{H}(x_i) = \frac{H(p_i)}{\log} \in [0,1]$ is the normalized Shannon entropy of the regime probabilities at location $i$ and $K$ is the number of regimes. Thus, $w_i \approx 1$ in regime cores (near-dterministic assignment) and $w_i \approx 0$ near interfaces (mixed surface). We report $I_{core}$ as a complementary diagnostic to $I_F$ separating stable within-regime drivers from transition-sensitive effects concentrated along regime boundaries.

**Data and study regions**

To rigorously validate the ZeGNN framework across heterogeneous domains, we curated three empirical datasets representing physical-environmental, socioeconomic, and epidemiological systems. We selected each dataset to reflect a system under substantial structural stress, thereby providing a strong test of the model's ability to identify regime-dependent mechanisms. We also generated a controlled simulation dataset suite in which the ground truth burden and capacity surfaces are known, so that mechanistic recovery can be evaluated under fully specified generating processes. The simulation domain is a 50 by 50 lattice with 2,500 locations and five spatially correlated covariates. We considered three scenarios called Global Linear, Local Linear, and Non-linear. Global Linear uses one regime, while Local Linear and the Non-linear scenario use three regimes generated by a Voronoi tessellation, which produces spatially contiguous but irregular regime boundaries.

To evaluate the framework's applicability to socioeconomic valuation dynamics, we utilized the King County housing dataset (May 2014–May 2015) provided by the GeoDa Center for Geospatial Analysis and Computation (Anselin, 2020). Grounded in a thermodynamic interpretation of Hedonic Pricing Theory (Rosen, 1974), we conceptualized housing price formation as a balance between quantitative load and qualitative resilience. Physical attributes quantifying the structural magnitude of the asset, such as square footage, number of bedrooms, bathrooms, and floors, were mapped to Burden ($E$), representing the baseline mass or maintenance load of the property. In contrast, qualitative attributes that sustain value against market depreciation or create scarcity premiums, including waterfront status, view, construction grade, and condition, were classified as Capacity ($S$), reflecting the asset's ability to maintain high-value regimes (Can, 1992; Bourassa et al., 2007).

We extended the framework to complex coupled social-environmental systems by analyzing neighborhood-level mental health prevalence. Following the Social Determinants of Health (SDOH) framework (Marmot, 2005) and the Stress-Buffering Hypothesis (Cohen & Wills, 1985), we modeled mental health outcomes as the emergent interplay between cumulative environmental stressors and community resources. Variables representing social deprivation and disorder, including neighborhood disadvantage index, crowding, rent burden, and density of alcohol outlets (beer, casinos), were encoded as burden ($E$), acting as cumulative physiological and psychosocial stressors (Gong et al., 2016). To offset this load, factors providing restorative environments or institutional support, such as nature parks (psychological restoration) and general hospitals (access to care), were classified as Capacity ($S$), capturing the community's potential to buffer against structural disadvantage (Barton & Rogerson, 2017; Robinette et al., 2021). By unifying these diverse domains under a shared burden-capacity formalism, we aim to demonstrate that the ZeGNN can diagnose universal topological features of regime shifts, whether driven by atmospheric blocking, market segmentation, or social inequality.

For the physical system, we analyzed fine particulate matter (PM2.5) concentrations across the Contiguous United States during June 2023, a period defined by the record-breaking Canadian wildfires that triggered severe air quality episodes across North America (Jain et al., 2024). This temporal window offers a unique "natural experiment" to evaluate the thermodynamic framework, as the atmosphere was subjected to an exogenous shock driven by specific synoptic anomalies. Notably, persistent blocking high-pressure systems over Canada that facilitated extreme fire weather, coupled with low-pressure troughs that channeled smoke southward into the Northeastern US (He et al., 2025; Zhang et al., 2024). Utilizing ERA5 reanalysis fields, we partitioned meteorological predictors into Burden ($E$) and Capacity ($S$) based on atmospheric chemistry and physics principles. Variables that thermodynamically favor pollution accumulation or secondary formation, specifically, 2m temperature and surface solar radiation (accelerating photochemical oxidation), and mean sea-level pressure (proxying synoptic stagnation and blocking), were classified as Burden (Seinfeld & Pandis, 2016; Jacob & Winner, 2009). Conversely, variables representing the atmosphere's kinetic energy to dilute or scavenge pollutants were mapped to Capacity; these include total precipitation (wet deposition), boundary layer height (vertical mixing volume), and wind components (horizontal ventilation) (Andronache, 2004; Miao et al., 2015).

**Baseline models**

To benchmark the Zentropy–GNN, we compare it against seven widely used baseline models spanning global and local spatial regression, tree ensembles, and generic neural architectures (Fotheringham et al., 2009, 2015; Dambon et al., 2021; Ni et al., 2022; De Sabbata &

Liu, 2023; Yin et al., 2024). All baseline models are trained on standardized covariates using the same 5-fold cross-validation splits and train–test partitions as the Zentropy–GNN, and we report performance on held-out data:

- Global Ordinary Least Squares (OLS): A conventional ordinary least squares model with an intercept, serving as a non-spatial reference that ignores spatial dependence but captures global linear trends.
- Geographically Weighted Regression (GWR): A spatially varying-coefficient linear model estimated with the fastGWR library (Li et al., 2019). For each outcome, we use an adaptive bisquare kernel with bandwidth (number of neighbors) selected by minimizing AICc by golden-section search, following standard practice for GWR. This provides a strong parametric spatial baseline that allows coefficients to vary smoothly across space.
- Decision Tree (DT): A non-parametric regression tree with maximum depth 10, at least 20 samples required to split an internal node, and a minimum of 10 samples per leaf. This configuration balances flexibility with regularization and avoids overly deep trees that overfit local idiosyncrasies.
- Random Forest (RF): an ensemble of regression trees rained with bootstrap aggregation. We use 300 trees with maximum depth 15, minimum 10 samples to split and 5 samples per leaf, whereas for the monthly PM2.5 experiment we use a computationally lighter configuration with 100 trees and the same depth and leaf constraints. In all cases we enable parallel training (all CPU cores) and use the default feature subsampling strategy.
- GBR (gradient boosting regressor): A gradient-boosted ensemble of shallow trees. We employ 100 estimators with learning rate of 0.1 and maximum depth 5; for PM2.5 we again use 100 estimators but with a smaller learning rate (0.05) and maximum depth 3 to favor smoother functions and reduce overfitting on smaller monthly samples.
- DNN (multilayer perceptron): A fully connected feed-forward neural network with two hidden layers of 64 units each and ReLU activations, trained on standardized inputs and outputs. The model is optimized with Adam (learning rate 0.001) using full-batch gradient updates for 600 hundred epochs, with gradient clipping (maximum norm 1.0) to stabilize training.
- Graph Neural Network (GNN): A generic graph neural network that operates on the same k-nearest-neighbor graph as the ZeGNN but without any thermodynamic decomposition. We construct a row-normalized adjacency matrix from the 12 nearest neighbors in great-circle distance and apply two message-passing steps interleaved with linear layers (64 hidden units) and ReLU activations, followed by a linear readout layer. The model is trained with Adam (learning rate 0.001) for 600 epochs with gradient clipping (maximum norm 1.0), using standardized covariates and the same spatial graph across all outcomes.

Across all experiments, hyperparameters for the tree-based models were selected via preliminary grid search on each outcome and then held fixed for the reported runs, while the neural baselines (GNN) share a common architecture and optimizer configuration to provide a fair comparison to ZeGNN.

**Acknowledgement:** This work was in part supported by the National Institutes of Health [grant number R01AI174892].

Supplementary Information

# Thermodynamic-Inspired Explainable GeoAI: Uncovering Regime-Dependent Mechanisms in Heterogeneous Spatial Systems

Sooyoung Lim[1], Zhenlong Li[1*], Zi-Kui Liu[2],

[1]Geoinformation and Big Data Research Lab, Department of Geography, Pennsylvania State University, University Park, PA 16801, USA
[2]Department of Materials Science and Engineering, Pennsylvania State University, University Park, PA 16801, USA
Corresponding Author*: Zhenlong Li(zhenlong@psu.edu)

# Supplementary Note 1: Configuring a Simulation Experiment

We designed a controlled spatial simulation experiment to rigorously evaluate the ZeGNN framework's capability to disentangle heterogeneous thermodynamic mechanisms. Specifically, this experiment aims to verify whether the model can (1) identify latent regimes defined by distinct generating functions rather than mere variable clustering, (2) correctly recover "role reversals" where the functional effect of a predictor flips sign across spatial domains, and (3) effectively internalize spatial autocorrelation relative to standard baselines. We consider three simulation scenarios (Global Linear, Local Linear, and Non-Linear), each sharing the same spatial domain and covariate fields but differing in how the thermodynamic potentials $E$ and $S$ are generated. In Supplementary Note 1, Panel A presents results from the Global linear scenario and Panel B presents results from the Local linear scenario. Results for the Nonlinear scenario are reported in the main manuscript to support the primary narrative. This note is organized in two parts. Section 1 describes simulation dataset generation. Section 2 reports simulation dataset results and indicates where each scenario is shown.

### 1.1 Simulation datasets generation

#### 1.1.1 Spatial Domain and Latent Regimes

The simulation domain is defined as a regular lattice (50×50; $N = 2{,}500$ locations) over the unit square $[0,1]^2$, with small coordinate jitter added to avoid distance ties and overly regular boundaries. For Global Linear, we set $K_{true} = 1$, so the entire domain belongs to a single regime. For Local Linear, we use a deterministic three-region partition consisting of an upper region, a lower-left region, and a lower-right region. For the Non-linear dataset, we generated $K_{true} = 3$ ground-truth regimes using a Voronoi tessellation based on three random seed points. These latent regimes serve as the structural containers for the distinct thermodynamic laws defined below.

#### 1.1.2 Spatially Structured Covariates.

We synthesized $p = 5$ spatially correlated covariates $\mathbf{x}_i = (x_{i1}, \dots, x_{i5})$, to represent a diverse set of spatially structured predictors. To ensure realistic spatial dependence, base fields were generated using Gaussian random fields (via KNN smoothing with $k = 15$) and subsequently standardized. Covariate generation is held fixed across the Global Linear, Local Linear, and Non-linear datasets; the differences across three datasets arise from how $E$ and $S$ are defined as functions of these covariates, not from changing the covariate fields themselves.

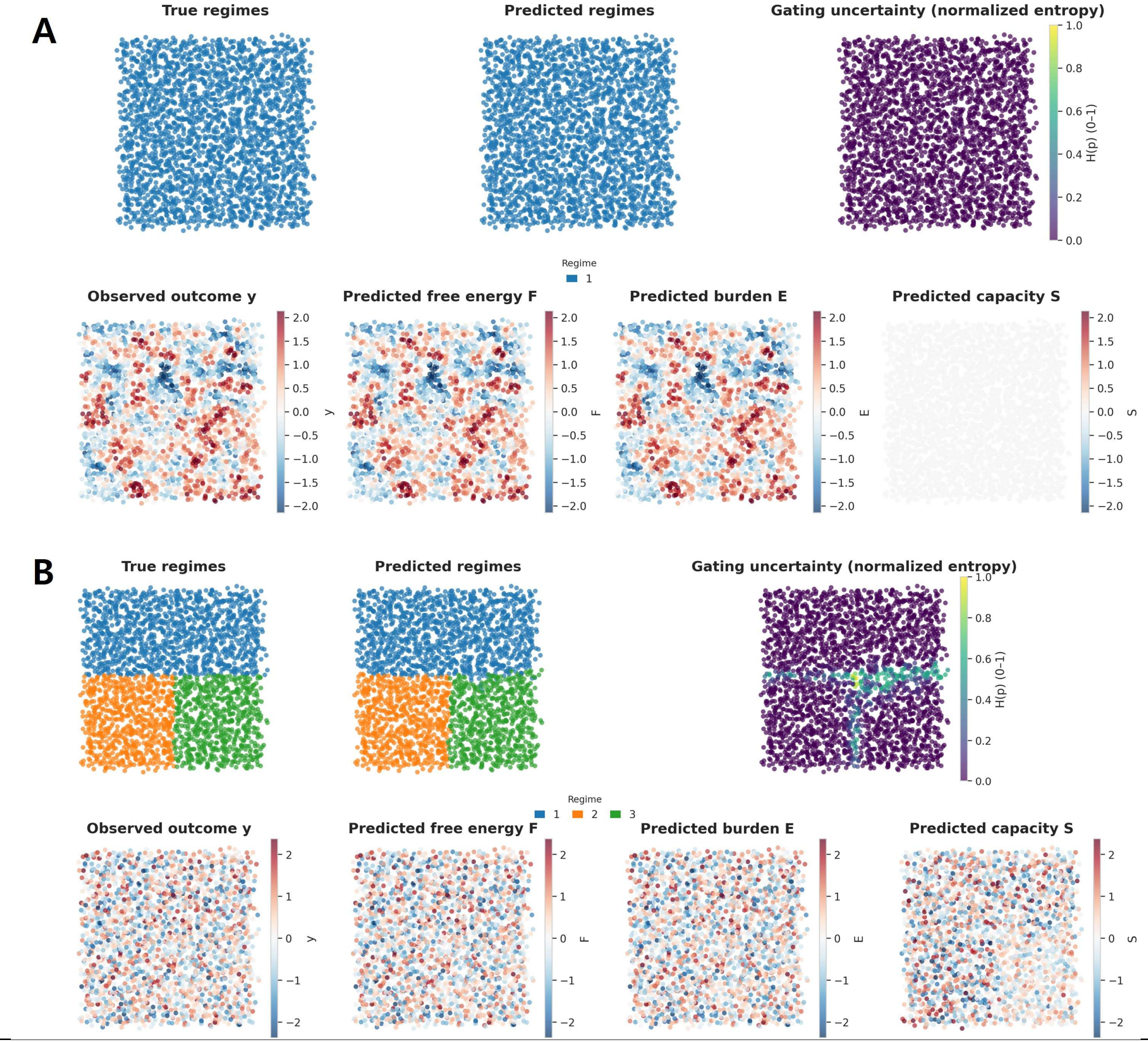

**Supplementary Fig. S1 | In sample simulation results for regime identification and thermodynamic decomposition.**

Panel A shows the Global linear scenario and Panel B shows the Local linear scenario. The nonlinear scenario is shown in the main manuscript. Within each panel, the top row shows the true regimes, the predicted regimes, and the gating uncertainty measured by normalized entropy. The bottom row shows the observed outcome y and the corresponding predicted free energy $F$, predicted burden $E$, and predicted capacity $S$, all displayed in standardized units for visual comparison.

### 1.1.3 Generative Mechanisms: Role Reversals and Structural Shifts

The core of this experiment lies in the regime-specific definitions of the Burden ($E$) and Capacity ($S$) potentials. Unlike standard simulations that vary only coefficients magnitudes, we injected qualitative structural shifts to test the model's mechanistic fidelity. The latent free energy is defined as $F(x) = E(x) - TS(x)$.

For a location $i$ in Regime 1 (Baseline State), the mechanism follows a standard linear-nonlinear mix where $x_1$ act as a positive burden driver:

$$E_i = 2.6x_{i1} + 2.2\tanh(2x_{i2}) + 1.6(x_{i3}^2 - 1)$$

$$S_i = 2(\sigma(2.5x_{i4}) - 0.5) + 2(x_{i5}\exp(-0.8x_{i5}^2))$$

Where $\sigma$ is the sigmoid function. Note the positive coefficient $+2.6$ for $x_1$.

In Regime 2 (Role Reversal State), the system shifts to a distinct thermodynamic state where the role of $x_1$ is explicitly inverted:

$$E_i = -2.6x_{i1} + 2(x_{i2}^2 - 1) + 1.8\tanh(2.2x_{i3})$$
$$S_i = 0.8\sin(3x_{i4}) + 0.9(\sigma(2x_{i5}) - 0.5)$$

Crucially, $x_1$ has a negative coefficient -2.6. This tests the model's ability to detect sign transitions without smoothing them out.

Finally, Regime 3 (Interaction State) represents a complex state driven by variable interactions and high-frequency oscillations:

$$E_i = 0.9x_{i1} + 2.6(x_{i2} \cdot x_{i3})$$
$$S_i = 1.7\sin(3.2x_{i4}) + 1.5(x_{i5}^2 - 1)$$

For ground-truth generation, we set the effective temperature $T = 1$ across all regimes by absorbing any temperature variations directly into the scaling of the Capacity ($S$) functions. The final observed outcome $y_i$ incorporates spatial diffusion and noise:

$$y_i = (E_i - S_i) + \rho \sum_{j \in N(i)} w_{ij} F_j + \eta_{spatial} + \epsilon_i$$

Where $\rho = 0.1$ represents the strength of spatial spillover, $\eta_{spatial}$ is an unobserved spatially structured error term, and $\epsilon_i \sim N(0, 0.12^2)$ is white noise.

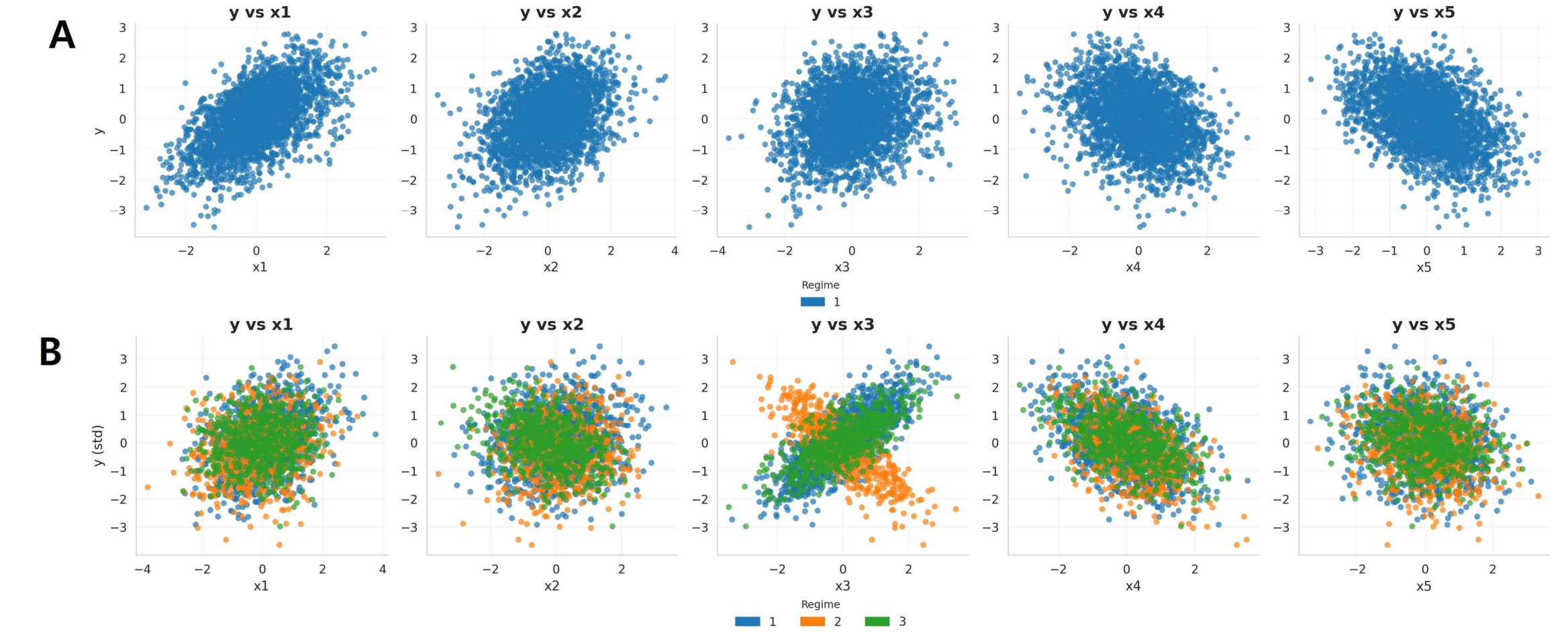

**Supplementary Fig. S2 | Outcome-covariate relationships by Scenario and regime**
Panel A shows the Global linear scenario and Panel B shows the Local linear scenario. The nonlinear scenario is shown in the main manuscript. Scatterplots of the standardized outcome $y$ against each standardized covariate $(x_1, \ldots, x_5)$ colored by the true latent regime. The panels visualize regime-dependent nonlinear response patterns and overlaps in feature space, motivating the need for regime-aware modeling rather than a single global mapping from $X$ to $y$.

## 1.2 Simulation datasets results

### 1.2.1 Validation by Gradient Matching

Because the data generating process relies on closed-form functions, we can derive the exact ground-truth thermodynamic sensitivities ($\nabla F_{true} = \frac{\partial E_{true}}{\partial x_j} - \frac{\partial S_{true}}{\partial x_j}$) at every location. This enables a rigorous gradient matching validation that compares the analytical ground truth against the model-learned gradients ($\nabla F_{pred}$).

For the linear driver $x_1$, the ground-truth partial derivative takes the form of a step function, shifting abruptly across regime boundaries:

$$\frac{\partial F_{true}}{\partial x_1} = \begin{cases} +2.6 & \text{in Regime 1} \\ -2.6 & \text{in Regime 2} \\ +0.9 & \text{in Regime 3} \end{cases}$$

This distinct sign flip serves as a strict test of the model's ability to detect role reversals without over-smoothing the transitions.

In contrast, for nonlinear drivers such as $x_2$, the partial derivative varies continuously as a function of the local state. For instance, across Regimes 1 and 2, the gradient follows distinct functional forms:

$$\frac{\partial F_{true}}{\partial x_2} = \begin{cases} 4.4\,\mathrm{sech}^2(2x_2) & \text{in Regime 1} \\ 4x_2 & \text{in Regime 2} \end{cases}$$

We quantify mechanistic fidelity by computing the spatial correlation between these analytical gradients and the model's learned sensitivities derived via automatic differentiation. High correlation confirms that ZeGNN has correctly identified the underlying physical laws, capturing both discrete regime shifts and continuous nonlinear dynamics.

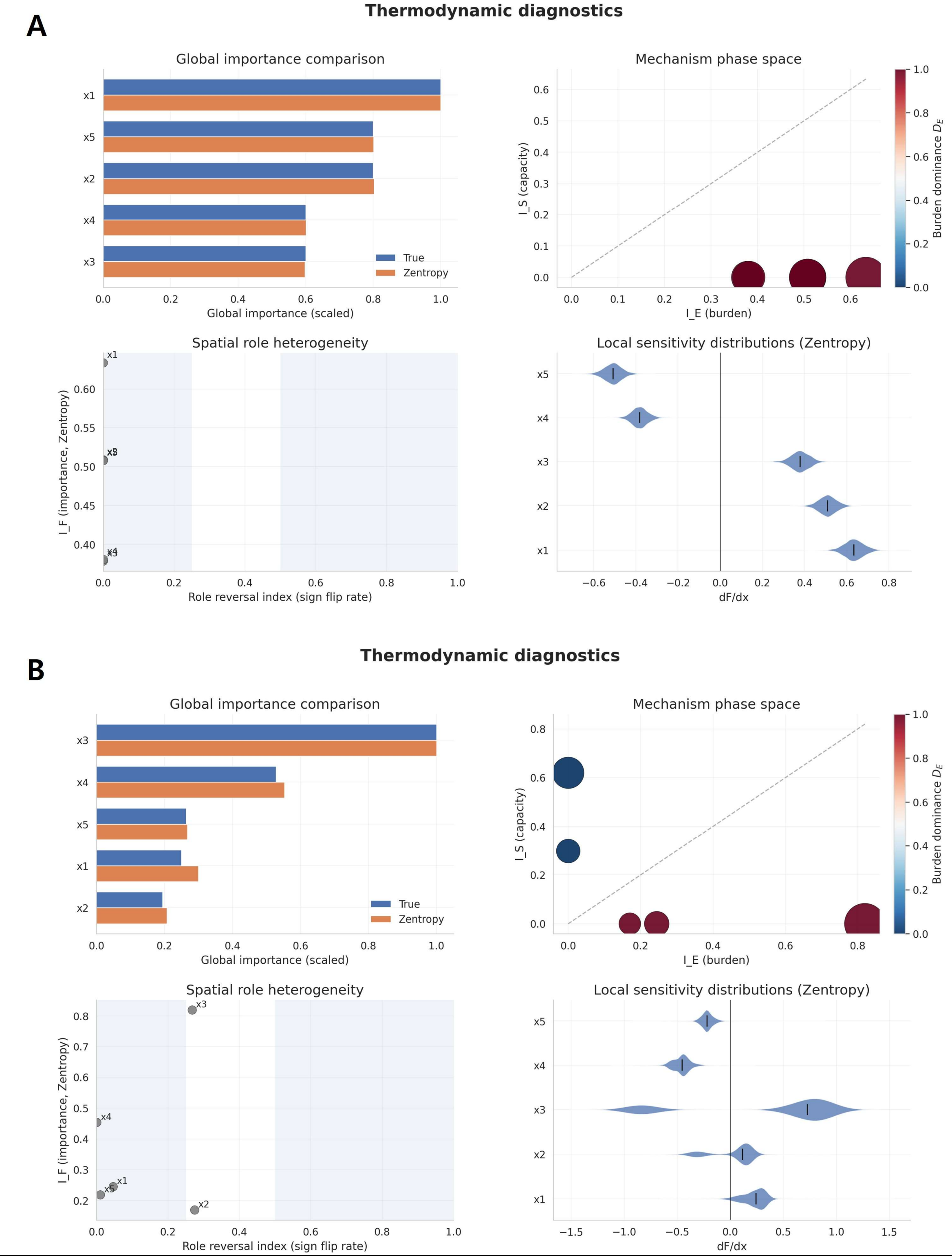

**Supplementary Fig. S3 | Thermodynamic mechanism diagnostics from gradient-based sensitivities**

Panel A shows the Global linear scenario and Panel B shows the Local linear scenario. The nonlinear scenario is shown in the main manuscript. Mechanism-level summaries derived from local gradient fields. Global feature importance summarized by the median magnitude of $|\partial F/\partial x_j|$. Burden-capacity phase space summarizing whether each covariate acts primarily through the burden pathway $E$ or the capacity pathway $S$ using channel-specific gradient magnitudes. Spatial role heterogeneity quantified by a role-reversal or sign-flip rate indicating whether a covariate's local effect changes sign across space/regimes. Distributions of local sensitivities $\partial F/\partial x_j$ highlighting heterogeneity and asymmetry in marginal effects.

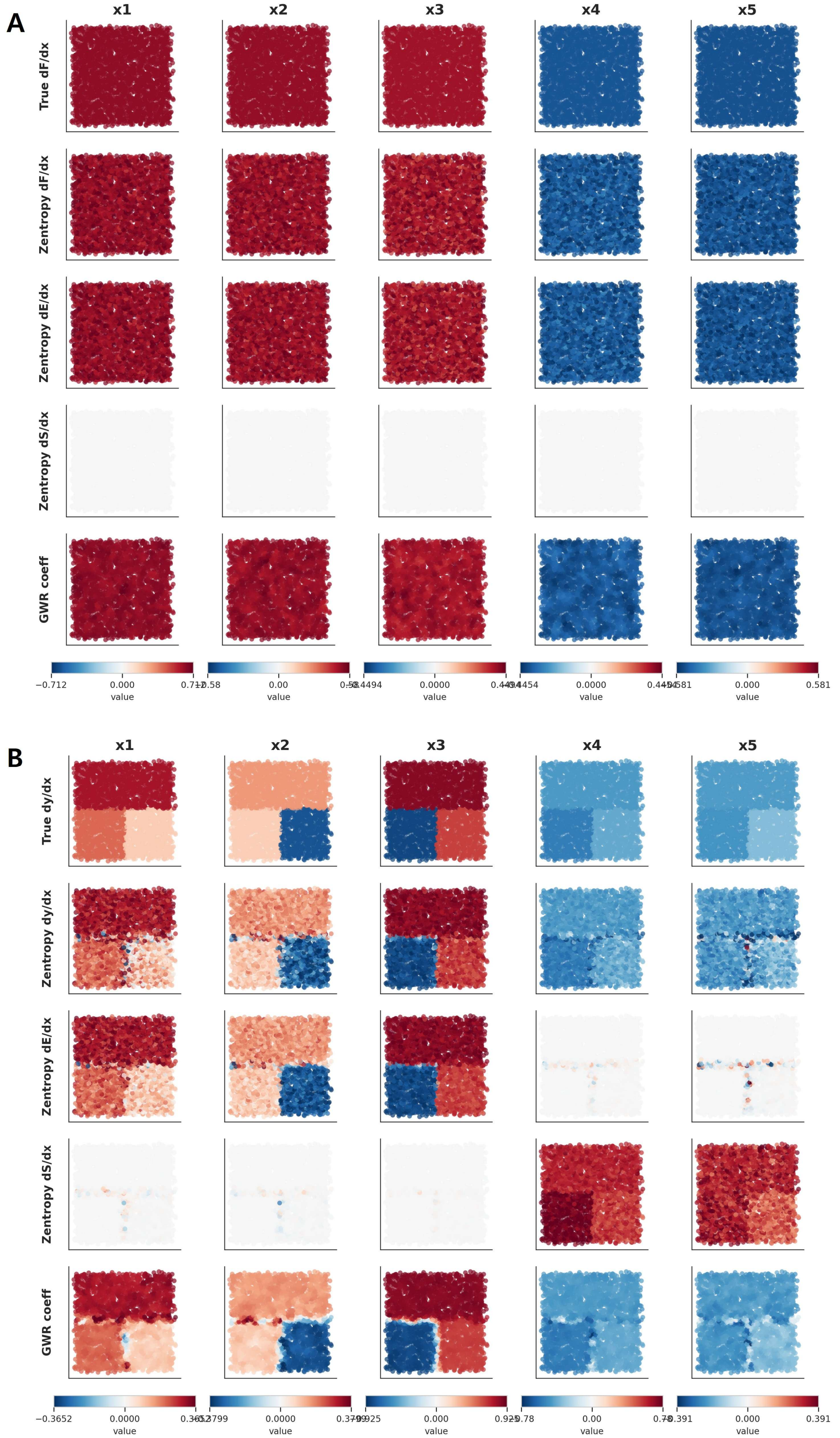

**Supplementary Fig. S4 | Local sensitivity decomposition map (true vs. predicted)**
Full spatial maps comparing ground-truth and model-recovered local sensitivities for all covariates. Rows report sensitivities of the free-energy output $\partial F/\partial x_j$ and its decomposition into burden and capacity pathways $\partial E/\partial x_j$ and $\partial S/\partial x_j$. Columns correspond to covariates $(x_1, \ldots, x_5)$. This expanded atlas complements the summary diagnostics in Supplementary Fig. S3 by showing where the model captures (or fails to capture) regime boundaries, transition corridors, and sign-reversal structures at the map level.

### 1.2.2 Preprocessing, graph construction, and evaluation protocols

For model fitting, we standardized all covariates and the outcome using z-score. Graph-based models were built on an undirected KNN graph using Euclidean distances in coordinate space, with graph size selected by blocked cross-validation from scenario-specific candidate sets. In the current simulation implementation, each simulation settings use $k \in \{8,10,12,14,16\}$. Residual spatial autocorrelation was quantified using Moran's' I computed on a KNN weights matrix (KNN size =8) for consistent cross-model comparison.
We evaluated models under two protocols: random 5-fold cross-validation and spatially blocked 5-fold cross-validation using a 5x5 grid to reduce spatial leakage. Predictive performance is summarized by fold-average $R^2$ and RMSE, and spatial fidelity is summarized by residual Moran's I. In the ZeGNN hyperparameter search, spatially blocked cross-validation serves as the primary selection protocol.

### 1.2.3 Preventing information leakage for graph-based models

Graph models can unintentionally incorporate test-fold information through message passing if edges linking to test nodes are retained during training. To prevent this, for each CV split we removed edges incident to test nodes and trained graph models only on the resulting training-node subgraph. Early stopping was conducted using an inner validation split drawn exclusively from training nodes (a fixed fraction of the training set), ensuring that test-fold labels and neighborhoods did not influence training dynamics.

## Supplementary Note 2: Data Description, Hyperparameter selection, and spatial diagnostics

| Variable | Role | Description |
|---|---|---|
| price | Outcome (y) | Sale price (USD). |
| long | Coordinate | Longitude (decimal degrees). |
| lat | Coordinate | Latitude (decimal degrees). |
| bedrooms | Burden (E) | Number of bedrooms. |
| bathrooms | Burden (E) | Number of bathrooms. |
| sqft_living | Burden (E) | Size of living area (square feet). |
| sqft_lot | Burden (E) | Size of lot (square feet). |
| floors | Burden (E) | Number of floors. |
| sqft_above | Burden (E) | Square feet above ground. |
| sqft_basement | Burden (E) | Square feet below ground (basement). |
| sqft_living15 | Burden (E) | Average living area of nearest 15 houses (square feet). |
| sqft_lot15 | Burden (E) | Average lot size of nearest 15 houses (square feet). |
| waterfront | Capacity (S) | Indicator: 1 if waterfront, 0 otherwise. |
| view | Capacity (S) | View quality index (0–4). |
| condition | Capacity (S) | House condition rating (1–5). |
| grade | Capacity (S) | Construction grade (materials/workmanship quality). |
| yr_built | Capacity (S) | Year built. |
| yr_renovated | Capacity (S) | Year renovated (0 if never renovated). |

**Table S1. Housing Dataset Description**
Variable definitions follow the GeoDa kingCounty House Sales documentation. Continuous predictors were used as recorded in the source dataset and (where applicable) standardized prior to model fitting (see Methods). yr_renovated = 0 indicates no renovation; waterfront is binary (0/1). Coordinates are provided in decimal degrees and were used for spatial graph construction after projection in analysis.

| Variable | Role | Description |
|---|---|---|
| Mental_health_prevelence | Outcome (y) | Prevalence (% or rate) of adults reporting poor mental health (commonly defined via "≥14 mentally unhealthy days in past 30 days"; small-area estimates). |
| X | Coordinate | Projected/planar X coordinate for tract centroid (units depend on CRS used in your preprocessing). |
| Y | Coordinate | Projected/planar Y coordinate for tract centroid (units depend on CRS used in your preprocessing). |
| PopulationDensity | Burden (E) | Population density (ACS-derived; tract-level). |
| PopulationFemale | Burden (E) | Female population (count or %; ACS-derived; tract-level). |
| Population65 | Burden (E) | Population aged 65+ (count or %; ACS-derived; tract-level). |
| NoHealthInsurance | Burden (E) | Population without health insurance (count or %; ACS-derived; tract-level). |
| Neighbor_Disadvantage | Burden (E) | Neighborhood disadvantage index (constructed composite; e.g., from ACS socioeconomic indicators). |
| Beer | Burden (E) | POI visitation/density proxy for alcohol-related venues (mobile phone place visitation aggregated to tract). |
| Casinos | Burden (E) | POI visitation/density proxy for casinos/gambling-related venues (tract-level). |
| DrinkingPlaces | Burden (E) | POI visitation/density proxy for drinking places (alcoholic beverages) (tract-level). |
| X.HouseRent | Burden (E) | House rent metric (ACS-derived; tract-level; exact definition depends on your preprocessing—e.g., median gross rent). |
| Generalhosptials | Capacity (S) | POI visitation/density proxy for general hospitals (health service utilization; tract-level). |
| NatureParks | Capacity (S) | POI visitation/density proxy for parks/nature access (positive health behavior; tract-level). |
| MentalHealthPati | Capacity (S) | POI visitation/density proxy for mental-health-related facilities/services (health service utilization; tract-level). |

**Table S2. Mental Health Dataset Description.**
Mental_health_prevelence represents the CDC frequent mental distress/poor mental health concept, commonly defined as reporting $\geq 14$ mentally unhealthy days in the past 30days (prevalence, % or rate; small-area estimates such as PLACES). Sociodemographic variables were compiled from ACS 5-year estimates at the census-tract level. POI-related variables are tract-level exposure proxies derived from mobile-device place activity/visitation datasets, consistent with the referenced Cities study. Variable names are preserved exactly as in the input files.

| Variable | Role | Description (ERA5; units) |
|---|---|---|
| pm | Outcome (y) | PM2.5 concentration field for June 2023. |
| x | Coordinate | Grid/centroid X coordinate (units depend on CRS). |
| y | Coordinate | Grid/centroid Y coordinate (units depend on CRS). |
| 2m temperature | Burden (E) | 2 m air temperature (K). |
| 2m dewpoint temperature | Burden (E) | 2 m dewpoint temperature (K). |
| Surface solar radiation downwards | Burden (E) | Surface short-wave (solar) radiation downwards, ssrd **(**J $m^{-2}$**;** accumulated over the time step / aggregated per product). |
| Mean sea level pressure | Burden (E) | Mean sea level pressure, msl (Pa). |
| Total precipitation | Capacity (S) | Total precipitation, tp **(**m; accumulated). |
| Boundary layer height | Capacity (S) | Boundary layer height, blh **(**m**).** |
| 10m u-component of wind | Capacity (S) | 10 m U wind component, 10u **(**m $s^{-1}$**).** |
| 10m v-component of wind | Capacity (S) | 10 m V wind component, 10v **(**m $s^{-1}$). |

**Table S3. PM2.5 Dataset Description**
Meteorological predictors are from ERA5 reanalysis. Units follow ECMWF conventions and aggregate to the temporal resolution used in this study. The analysis focuses on June 2023.

| case | model | $R^2$ (in sample) | $R^2$ (random cv) | $R^2$ (spatial cv) | RMSE (in sample) | RMSE (random cv) | RMSE (spatial cv) | Moran's I (in sample) |
|---|---|---|---|---|---|---|---|---|
| Housing | OLS | 0.654 | 0.651 | 0.566 | 216,039 | 217,015 | 241,784 | 0.470 |
| | GWR | 0.912 | 0.880 | 0.727 | 108,858 | 127,153 | 191,997 | 0.102 |
| | DT | 0.783 | 0.669 | 0.557 | 170,839 | 211,321 | 244,239 | 0.333 |
| | RF | 0.873 | 0.748 | 0.639 | 130,841 | 184,413 | 220,478 | 0.361 |
| | GBR | 0.848 | 0.757 | 0.659 | 143,033 | 181,096 | 214,538 | 0.471 |
| | DNN | 0.826 | 0.757 | 0.666 | 153,151 | 180,838 | 212,040 | 0.436 |
| | GNN | 0.611 | 0.572 | 0.501 | 228,956 | 240,056 | 259,468 | 0.082 |
| | ZeGNN | 0.941 | 0.896 | 0.838 | 89,436 | 118,665 | 147,639 | 0.070 |
| Mental Health | OLS | 0.612 | 0.603 | 0.453 | 2.195 | 2.218 | 2.604 | 0.626 |
| | GWR | 0.907 | 0.846 | 0.065 | 1.077 | 1.382 | 3.405 | 0.076 |
| | DT | 0.737 | 0.680 | 0.608 | 1.806 | 1.994 | 2.206 | 0.445 |
| | RF | 0.867 | 0.748 | 0.672 | 1.287 | 1.768 | 2.018 | 0.494 |
| | GBR | 0.773 | 0.748 | 0.673 | 1.676 | 1.767 | 2.015 | 0.517 |
| | DNN | 0.768 | 0.755 | 0.674 | 1.697 | 1.742 | 2.012 | 0.505 |
| | GNN | 0.545 | 0.536 | 0.394 | 2.376 | 2.399 | 2.742 | 0.238 |
| | ZeGNN | 0.889 | 0.873 | 0.770 | 1.174 | 1.254 | 1.690 | 0.225 |
| PM2.5 | OLS | 0.272 | 0.259 | -0.040 | 1.964 | 1.982 | 2.349 | 0.368 |
| | GWR | 0.550 | 0.445 | 0.053 | 1.544 | 1.716 | 2.241 | 0.111 |
| | DT | 0.664 | 0.366 | 0.071 | 1.335 | 1.834 | 2.220 | 0.069 |
| | RF | 0.787 | 0.518 | 0.242 | 1.063 | 1.600 | 2.005 | 0.027 |
| | GBR | 0.606 | 0.448 | 0.155 | 1.445 | 1.711 | 2.117 | 0.171 |
| | DNN | 0.797 | 0.453 | 0.206 | 1.038 | 1.704 | 2.052 | -0.004 |
| | GNN | 0.518 | 0.424 | -0.053 | 1.599 | 1.747 | 2.362 | 0.015 |
| | ZeGNN | 0.820 | 0.474 | 0.278 | 0.977 | 1.670 | 1.957 | -0.049 |

**Table S4. Comparative predictive performance and residual spatial autocorrelation across models and case studies.**
This table reports in-sample, random cross-validation, and spatial cross-validation performance for all benchmark models and ZeGNN across the Housing, Mental Health, and PM2.5 case studies. Performance is summarized using $R^2$ and RMSE, and residual spatial dependence is evaluated using in-sample Moran's I. The table provides the exact numerical values underlying the comparative patterns visualized in Fig. 3 and highlights the extent to which performance changes under spatially blocked evaluation.

| case | Graph (k) | K | $\lambda_{sparse}$ | $\lambda_{mag}$ | $R^2$ (Spatial) | RMSE (Spatial) | Residual Moran's I | $n_{eff,global}$ | $n_{eff,local}$ | Max dominant-regime share |
|---|---|---|---|---|---|---|---|---|---|---|
| Housing | 8 | 5 | 0.1 | 0.001 | 0.838 | 0.402 | 0.070 | 5.000 | 3.425 | 0.273 |
| Mental Health | 14 | 5 | 0.01 | 0.001 | 0.770 | 0.480 | 0.225 | 4.986 | 3.611 | 0.318 |
| PM2.5 | 12 | 5 | 0.001 | 0.0001 | 0.278 | 0.850 | -0.049 | 3.867 | 2.687 | 0.673 |

**Table S5. Selected hyperparameters and fitted diagnostics for the empirical case studies.**
For each case study, the table reports the selected graph neighborhood size $k$, regime upper bound $K_{upper}$, sparse regularization $\lambda_{sparse}$, and magnitude regularization $\lambda_{mag}$, together with spatial cross-validation performance and post-fir diagnostics. Residual Moran's I is reported for the refit in-sample residuals. $n_{eff,global}$ summarizes the effective number of regimes used globally, $n_{eff,local}$ summarizes the local effective number of regimes, and the maximum dominant-regime share reports the largest observed share of a single dominant regime across spatial units.

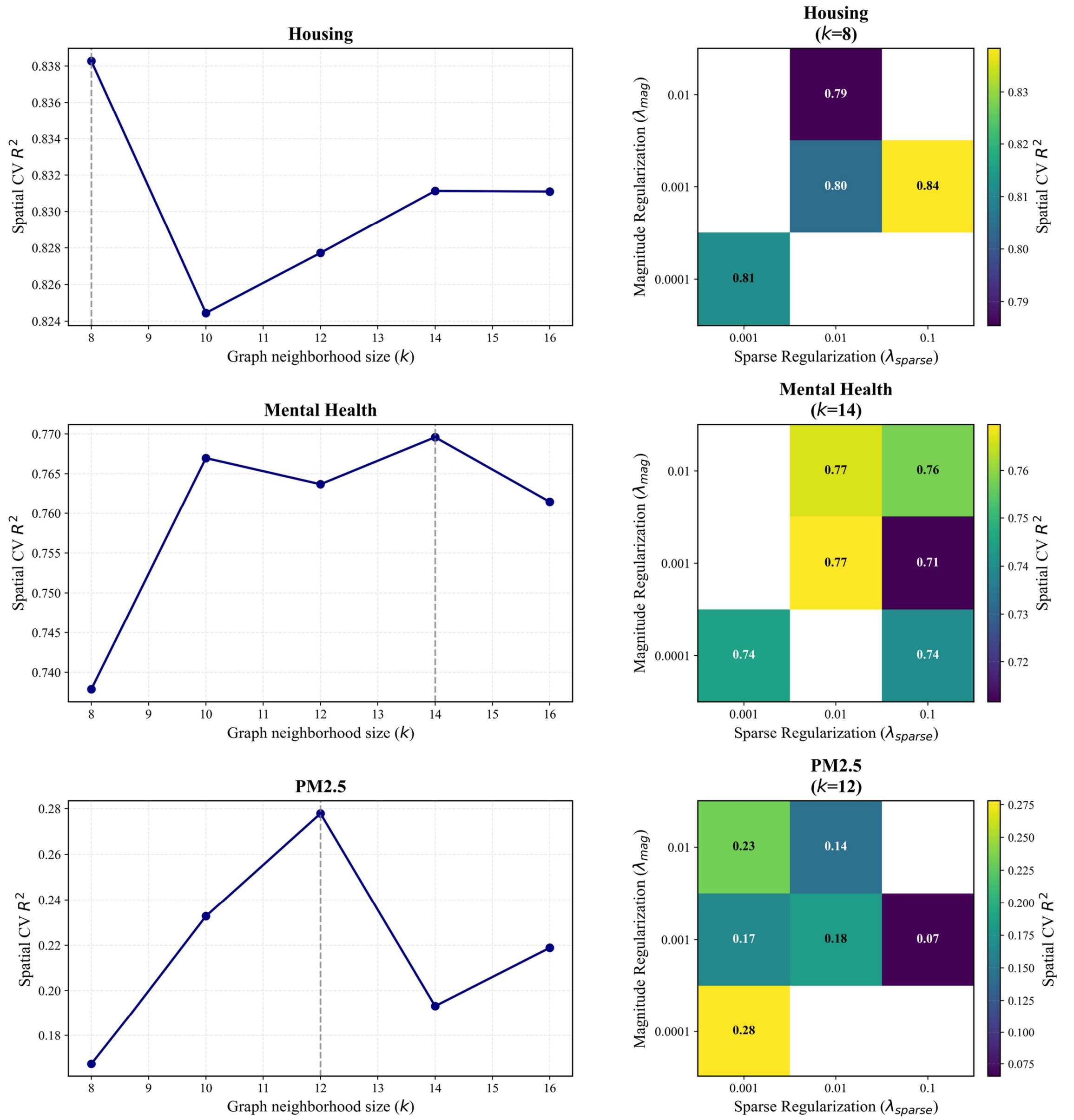

**Supplementary Fig. S5 | Spatial block cross-validation selection of graph neighborhood size and regularization**

Left panels show mean spatial cross-validation $R^2$ as a function of graph neighborhood size $k$ for each case study. Dashed vertical lines indicate the selected $k$ used in the final model for each case. Right panels show mean spatial cross-validation $R^2$ across the $\lambda_{sparse} \times \lambda_{mag}$ grid at the selected $k$. Titles report the selected $k$ for each case. Blank cells indicate configurations that were not evaluated or not retained in the sampled search.

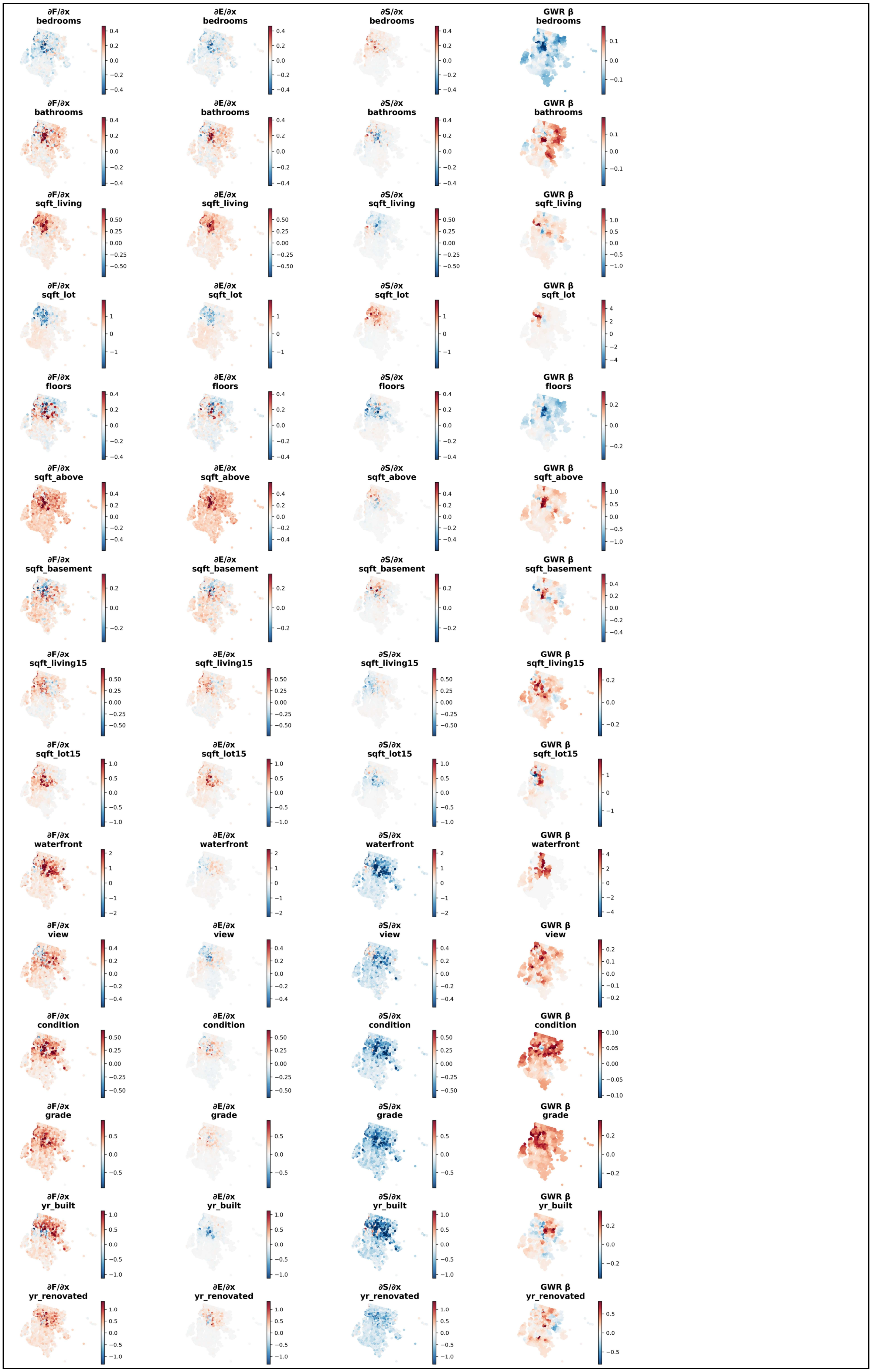

**Supplementary Fig. S6 | Housing thermodynamic sensitivity atlas and GWR comparison (local derivative decomposition).**
Local derivative-based sensitivity maps for the Housing case under configuration. For each predictor $x_j$, panels report spatial maps of the total free-energy sensitivity $\partial F/\partial x_j$, its decomposition into burden $\partial E/\partial x_j$, and capacity $\partial S/\partial x_j$ pathways, and the corresponding local coefficients from Geographically Weighted Regression (GWR) for comparison. Positive values indicate that increasing $x_j$ locally increases the corresponding thermodynamic component, whereas negative values indicate local decreases. These maps visualize where a covariate operates primarily through the burden pathway ($E$) versus the capacity pathway ($S$), while the GWR panels provide a baseline to contrast the thermodynamic regime structures against standard smooth spatial variations.

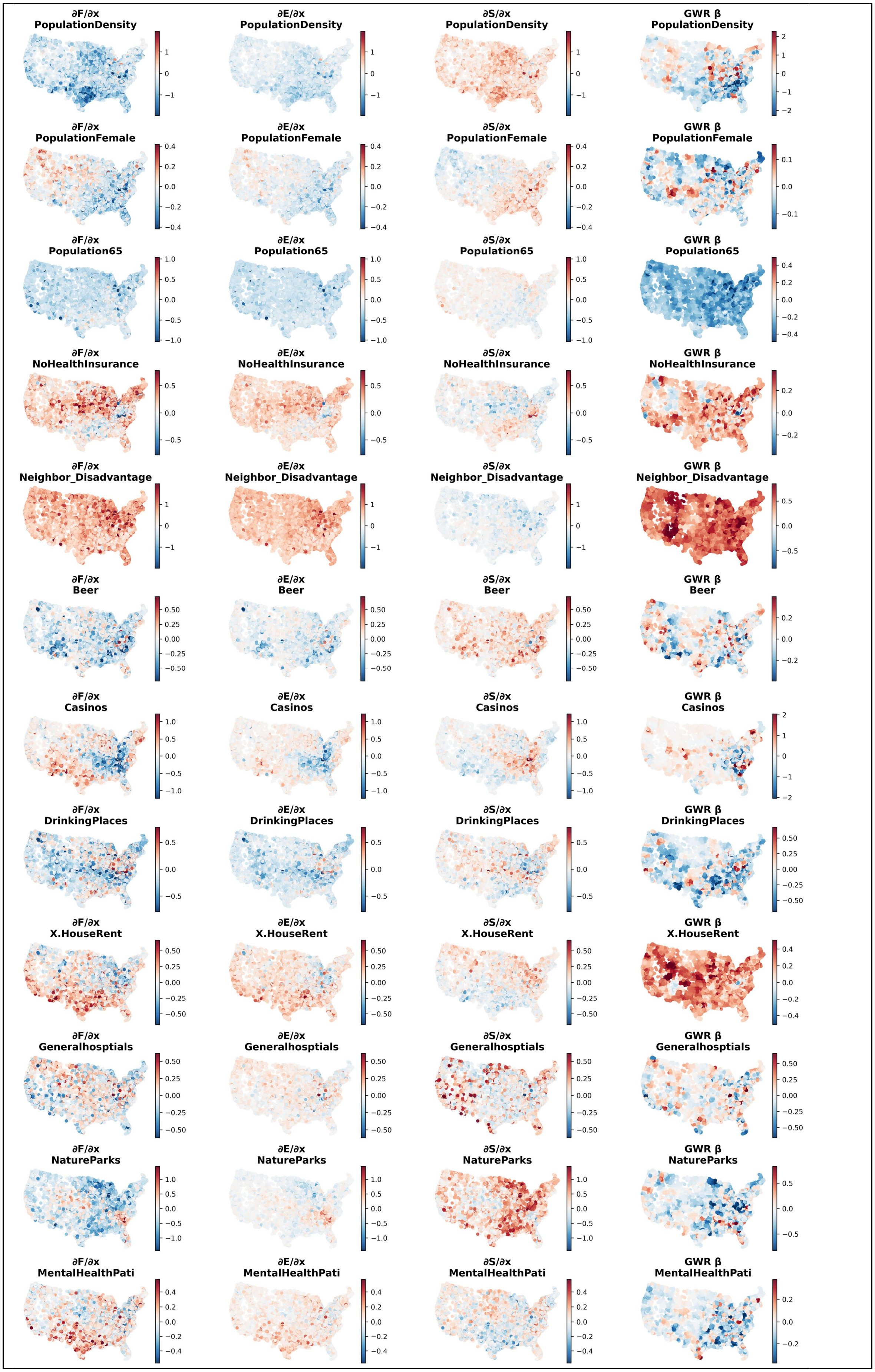

**Supplementary Fig. S7 | Mental health thermodynamic sensitivity atlas and GWR comparison (local derivative decomposition).**
Local derivative-based sensitivity maps for the Mental Health case under configuration. For each predictor $x_j$, panels report spatial maps of the total free-energy sensitivity $\partial F/\partial x_j$, its decomposition into burden $\partial E/\partial x_j$, and capacity $\partial S/\partial x_j$ pathways, and the corresponding local coefficients from Geographically Weighted Regression (GWR) for comparison. Positive values indicate that increasing $x_j$ locally increases the corresponding thermodynamic component, whereas negative values indicate local decreases. These maps visualize where a covariate operates primarily through the burden pathway ($E$) versus the capacity pathway ($S$), while the GWR panels provide a baseline to contrast the thermodynamic regime structures against standard smooth spatial variations.

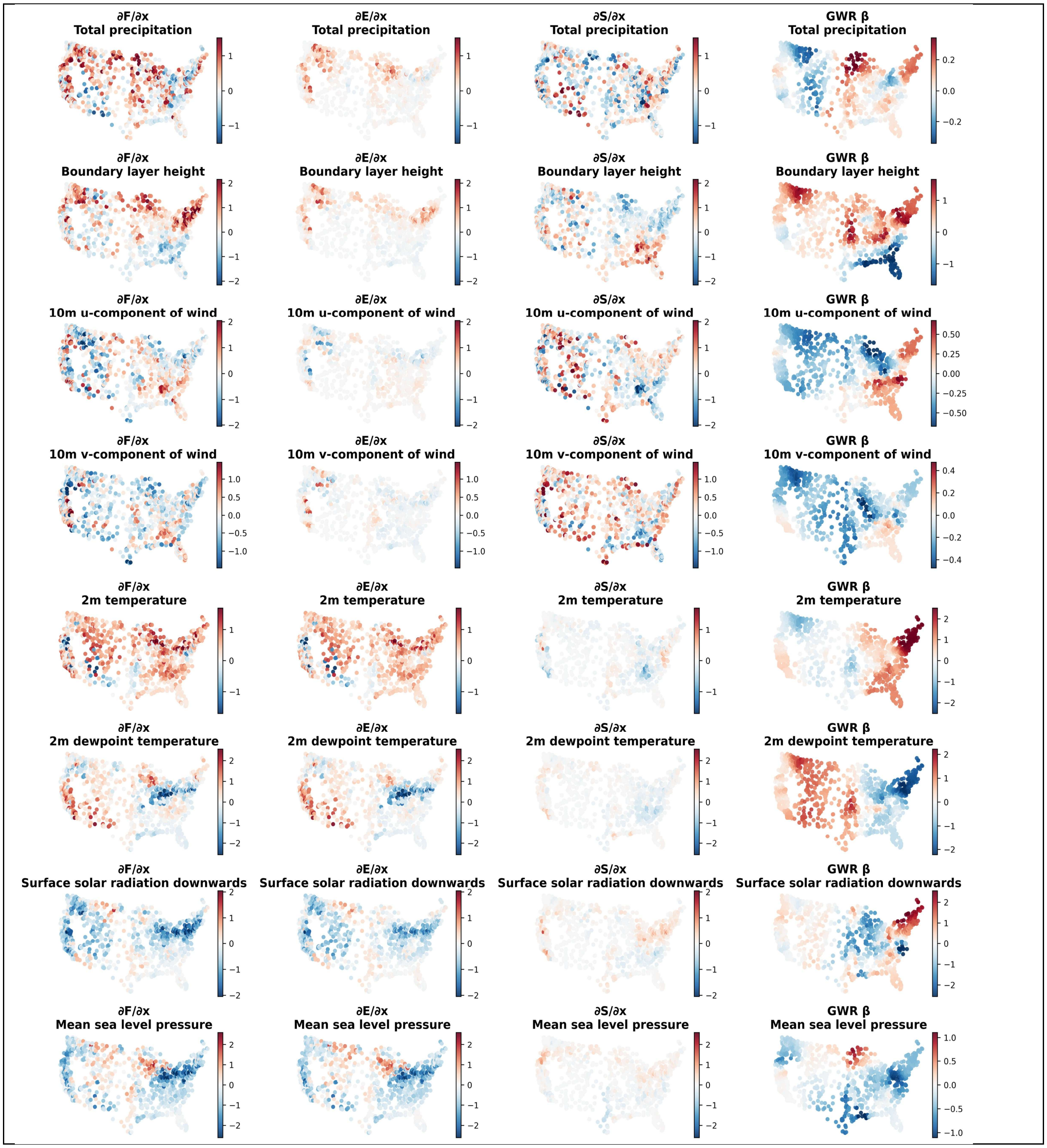

**Supplementary Fig. S8 | PM2.5 (June 2023) thermodynamic sensitivity atlas and GWR comparison (local derivative decomposition).**

For each predictor $x_j$, panels report spatial maps of the total free-energy sensitivity $\partial F/\partial x_j$, its decomposition into burden $\partial E/\partial x_j$, and capacity $\partial S/\partial x_j$ pathways, and the corresponding local coefficients from Geographically Weighted Regression (GWR) for comparison. Positive values indicate that increasing $x_j$ locally increases the corresponding thermodynamic component, whereas negative values indicate local decreases. These maps visualize where a covariate operates primarily through the burden pathway $(E)$ versus the capacity pathway $(S)$. Notably, the spatial footprint of the Burden sensitivity ($\partial E/\partial x_j$) closely aligns with the trajectory of the wildfire smoke plume, confirming that the model physically grounds the wildfire impact within the entropic potential term ($E$) rather than distributing it as random residual noise or smooth spatial trends as seen in GWR.